\documentclass[twoside]{article}

\usepackage[accepted]{aistats2019}

\usepackage[utf8]{inputenc} 
\usepackage[T1]{fontenc}    
\usepackage{hyperref}       
\usepackage{url}            
\usepackage{booktabs}       
\usepackage{amsfonts}       
\usepackage{nicefrac}       
\usepackage{microtype}      

\usepackage{amsmath,amscd,amssymb}
\usepackage{array}
\usepackage{bbold}
\usepackage{bm}
\usepackage{mathtools}
\usepackage{hyperref}

\usepackage[table]{xcolor}
\usepackage{graphicx}
\usepackage{subcaption}
\usepackage{tabu}
\usepackage[ruled]{algorithm2e}

\usepackage{scalerel,stackengine}
\stackMath

\newcommand{\bx}{{\bm{x}}}

\newcommand{\bth}{{\bm{\theta}}}
\newcommand{\beps}{{\bm{\epsilon}}}
\newcommand{\bz}{{\bm{z}}}
\newcommand{\bmu}{{\bm{\mu}}}
\newcommand{\bml}{{\bm{\lambda}}}

\newcommand{\E}{\mathbb{E}}
\newcommand*\VF[1]{\mathbf{#1}}
\newcommand*\dif{\mathop{}\!\mathrm{d}}

\begin{document}
\begin{NoHyper}

\renewcommand{\thefootnote}{\fnsymbol{footnote}}
\runningauthor{Martin Jankowiak, Theofanis Karaletsos}

\twocolumn[

\aistatstitle{Pathwise Derivatives for Multivariate Distributions}

\aistatsauthor{ Martin Jankowiak\footnotemark \And Theofanis Karaletsos}

\aistatsaddress{ Uber AI Labs, San Francisco, CA } ]

\footnotetext{Correspondence to:  jankowiak@uber.com}
\end{NoHyper}
\renewcommand{\thefootnote}{\arabic{footnote}}

\begin{abstract}
We exploit the link between the transport equation and derivatives of expectations to construct
efficient pathwise gradient estimators for multivariate distributions. We focus on two main threads.
First, we use null solutions of the transport equation to construct adaptive control variates that can
be used to construct gradient estimators with reduced variance. 
Second, we consider the case of multivariate mixture distributions. In particular we show how to compute pathwise
derivatives for mixtures of multivariate Normal distributions with arbitrary means and diagonal covariances. 
We demonstrate
in a variety of experiments in the context of variational inference that our gradient estimators can outperform other methods, 
especially in high dimensions.
\end{abstract}


\section{Introduction}

Stochastic optimization is a major component of many machine learning algorithms. 
In some cases---for example in variational inference---the stochasticity arises from 
an explicit, parameterized distribution $q_{\bth}(\bz)$. In these cases the optimization problem can often be cast as maximizing
an objective 
\begin{equation}
\label{eqn:obj}
\mathcal{L} = \E_{q_{\bth}(\bz)} \left[ f(\bz) \right]
\end{equation}
where $f(\bz)$ is a (differentiable) test function and $\bth$ is a vector of parameters.\footnote{Without loss of generality we assume that $f(\bz)$ has no explicit dependence on $\bth$, since gradients of $f(\bz)$ are readily handled by bringing $\nabla_\bth$ inside the expectation.}
 In order to maximize $\mathcal{L}$ we would like to follow gradients $\nabla_\bth \mathcal{L}$. 
If $\bz$ were a \emph{deterministic} function of $\bth$ we would simply apply the chain rule; for any 
component ${\theta_\alpha}$ of $\bth$ we have
\begin{equation}
\label{eqn:detf}
\nabla_{\theta_\alpha} f(\bz) = \nabla_\bz f \cdot \frac{\partial\bz}{\partial{\theta_\alpha}}
\end{equation}
What is the generalization of Eqn.~\ref{eqn:detf} to the stochastic case? As observed in \cite{beyond}, 
we can construct an unbiased estimator for the gradient of Eqn.~\ref{eqn:obj} provided we can solve the following partial differential
equation (the transport a.k.a.~continuity equation):
\begin{equation}
\label{eqn:transport}
\frac{\partial}{\partial {\theta_\alpha}} q_\bth + \nabla_\bz \cdot \left( q_\bth {\bm v}^{\theta_\alpha} \right)=0
\end{equation}
Here ${\bm v}^{\theta_\alpha}$ is a vector field defined on the sample space of $q_\bth(\bz)$.\footnote{More explicitly: 
at each point $\bz$ in the sample space the vector field ${\bm v}^{\theta_\alpha}$ is specified by $D$ components ${\bm v}^{\theta_\alpha}_i$,
$i=1,...,D$,
where $D$ is the dimension of the sample space.} Note that there is
a vector field ${\bm v}^{\theta_\alpha}$ for each component ${\theta_\alpha}$ of $\bth$. We can then form the following {\it pathwise} 
gradient estimator:
\begin{equation}
\label{eqn:estimator}
\nabla_{\theta_\alpha} \mathcal{L} = \E_{q_\bth(\bz)} \left[ \nabla_\bz f \cdot {\bm v}^{\theta_\alpha}  \right]
\end{equation}
Pathwise gradient estimators are particularly appealing because, empirically, they 
generally exhibit lower variance than alternatives. 

That the gradient estimator in Eqn.~\ref{eqn:estimator} is unbiased follows directly from the divergence theorem:
\begin{equation}
\begin{split}
\nonumber
\nabla_{\theta_\alpha} \mathcal{L}  &= \int d\bz  f(\bz) \frac{\partial q_\bth(\bz)}{\partial {\theta_\alpha}} 
=-\! \int d\bz f(\bz)  \nabla_\bz \cdot \left( q_\bth {\bm v}^{\theta_\alpha} \right) \\
&=  \int_V d\bz  \nabla_\bz  f \cdot (q_\bth{\bm v}^{\theta_\alpha}) -
\oint_{S} f q_\bth {\bm v}^\theta \cdot \hat{\VF{n}} \dif S\\
&= \lim_{S\to\infty} \left\{ \int_V d\bz   \nabla_\bz  f \cdot (q_\bth{\bm v}^{\theta_\alpha}) -
\oint_{S} f q_\bth {\bm v}^\theta \cdot \hat{\VF{n}} \dif S \right\} \\
&= \int d\bz  q_\bth(\bz) \nabla_\bz  f \cdot {\bm v}^{\theta_\alpha} 
=\E_{q_\bth(\bz)} \left[ \nabla_\bz f \cdot {\bm v}^{\theta_\alpha}  \right]
\end{split}
\end{equation}
Here we have substituted for $\frac{\partial}{\partial {\theta_\alpha}} q_\bth$ using Eqn.~\ref{eqn:transport} in the first line,
appealed to the divergence theorem\footnote{More explicitly: we apply the divergence theorem to the vector field $f q_\bth {\bm v}^\theta$ and expand the divergence in the volume integral using the product rule (cf.~Green's identities).} in the second line, and taken the surface to infinity in the third line. In the final step we assume that $f q_\bth{\bm v}^\theta $ is sufficiently well-behaved that we can drop the surface integral as the surface goes to infinity.

Thus Eqn.~\ref{eqn:estimator} is a generalization of Eqn.~\ref{eqn:detf} to the stochastic case, 
with the velocity field ${\bm v}^{\theta_\alpha}$ playing the role
of $\frac{\partial\bz}{\partial{\theta_\alpha}}$ in the deterministic case. In contrast to the deterministic case where $\frac{\partial\bz}{\partial{\theta_\alpha}}$ is
uniquely specified, Eqn.~\ref{eqn:transport} generally
admits an infinite dimensional space of solutions for ${\bm v}^{\theta_\alpha}$. 

The transport equation encodes an intuitive geometric picture. As we vary $\bth$ we move
$q_{\bth}(\bm{z})$ along a curve in the space of distributions over the sample space.  This curve corresponds to
a time-varying cloud of particles; in this analogy the set of
velocity fields $\{ \bm{v}^{\theta_\alpha} \}_\alpha$ describes the (infinitesimal)
displacements that particles undergo as $\bth$ is varied. Because each velocity field obeys the corresponding transport equation, Eqn.~\ref{eqn:transport},
the total probability is conserved and the displaced particles are distributed according to the displaced distribution.

In this work we are interested in the multivariate case, where the solution space to Eqn.~\ref{eqn:transport} is very rich, and where 
alternative gradient estimators tend to suffer from large variance, especially in high dimensions. 
Our contributions are two-fold. In Sec.~\ref{sec:avf} we exploit the the fact that
the transport equation admits a large space of solutions to construct gradient estimators that are \emph{adapted} to the test function $f(\bz)$. 
In the context of the reparameterization trick (see Sec.~\ref{sec:reptrick}) our approach is conceptually analogous to adaptively choosing
a reparameterization $\mathcal{T}$ from a family of transformations $\{ \mathcal{T}_\bml(\bm{\epsilon}; \bth) \}$, with the difference that the transformation is never explicitly constructed;
rather it is implicit as a solution to Eqn.~\ref{eqn:transport}. Second, in Sec.~\ref{sec:mix} we construct pathwise gradient estimators for mixtures
of Normal distributions with diagonal covariance matrices. We emphasize that the resulting gradient estimators are the only known 
unbiased pathwise gradient estimators for this class of mixture distributions whose computational complexity is linear in the dimension of the sample space.
 As we show in Sec.~\ref{sec:exp} our gradient estimators often exhibit low variance and good performance as compared
to alternative methods.

The rest of this paper is organized as follows. 
In Sec.~\ref{sec:background} we give an overview of stochastic gradient variational inference (SGVI) and stochastic gradient estimators.
In Sec.~\ref{sec:avf} we use parameterized families of solutions to Eqn.~\ref{eqn:transport} to construct adaptive pathwise gradient estimators. 
In Sec.~\ref{sec:mix} we present solutions to Eqn.~\ref{eqn:transport} for multivariate mixture distributions.
In Sec.~\ref{sec:related} we place our work in the context of recent research.
In Sec.~\ref{sec:exp} we validate our proposed techniques with a variety of experiments in the context of variational inference.
In Sec.~\ref{sec:discussion} we  discuss directions for future work.


\section{Background}
\label{sec:background}

\subsection{Stochastic Gradient Variational Inference}

Given a probabilistic model $p(\bx, \bz)$ with observations $\bx$ and latent random variables $\bz$, variational inference 
recasts posterior inference as an optimization problem. Specifically we define a family of variational distributions $q_\bth(\bz)$ parameterized by 
$\bth$ and seek to find a value of $\bth$ that minimizes the KL divergence between $q_\bth(\bz)$ and the (unknown) posterior 
$p(\bz|\bx)$ \cite{jordan1999introduction,paisley2012variational,hoffman2013stochastic,wingate2013automated,ranganath2014black}. 
This is equivalent to maximizing the ELBO, defined as
\begin{equation}
\begin{split}
{\rm ELBO} = \E_{q_\bth(\bz)} \left[ \log p(\bx,\bz) - \log q_\bth(\bz) \right]
\end{split}\label{eq:elbo}
\end{equation}
This stochastic optimization problem will be the basic setting for most of our experiments in Sec.~\ref{sec:exp}.

\subsection{Score Function Estimator}

The score function estimator, also referred to as \textsc{reinforce} \cite{glynn1990likelihood,williams1992simple,fu2006gradient}, provides a 
simple and broadly applicable recipe for estimating gradients, with the simplest variant given by
\begin{equation}
\nonumber
\nabla_{\theta_\alpha} \mathcal{L} =  \E_{q_{\bth}(\bz)} \left[ f(\bz) \nabla_{\theta_\alpha} \log q_{\bth}(\bz) \right]
\end{equation}
Although the score function estimator is very general (e.g.~it can be applied to discrete random variables) it
typically suffers from high variance, although this can be mitigated with the use of variance reduction techniques
such as Rao-Blackwellization \cite{casella1996rao} and control variates \cite{ross}.

\subsection{Reparameterization Trick}
\label{sec:reptrick}

The reparameterization trick (RT) is not as broadly applicable
as the score function estimator, but it generally exhibits lower variance 
\cite{price1958useful,salimans2013fixed,kingma2013auto,glasserman2013monte,titsias2014doubly,rezende2014stochastic}. 
It is applicable to continuous
random variables whose probability density $q_{\bth}(\bz)$ can be reparameterized such that we can 
rewrite expectations $\E_{q_{\bth}(\bz)} \left[ f(\bz) \right]$ as $\E_{q_0(\bm{\epsilon})} \left[ f(\mathcal{T}(\bm{\epsilon}; \bth)) \right ]$,
where $q_0(\bm{\epsilon})$ is a fixed distribution and $\mathcal{T}(\bm{\epsilon}; \bth)$ is a differentiable
$\bth$-dependent transformation. 
Since the expectation w.r.t.~$q_0(\bm{\epsilon})$ has no $\bth$ dependence, gradients w.r.t.~$\bth$
can be computed by pushing $\nabla_\bth$ through the expectation. 
This reparameterization can be done for a number of distributions, including for example the Normal distribution.
As noted in \cite{beyond}, in cases where the reparameterization trick is applicable, we have that
\begin{equation}
{\bm v}^{\theta_\alpha} = \frac{\partial \mathcal{T}(\beps; \bth)}{\partial {\theta_\alpha}}\bigg\rvert_{\beps=\mathcal{T}^{-1}(\bz; \bth)}
\end{equation}
is a solution to the transport equation, Eqn.~\ref{eqn:transport}. 


\section{Adaptive Velocity Fields}
\label{sec:avf}

Whenever we have a solution to Eqn.~\ref{eqn:transport} we can form an unbiased gradient estimator 
via Eqn.~\ref{eqn:estimator}. An intriguing possibility arises if we can construct a parameterized \emph{family} of
solutions $\bm{v}^{\theta_\alpha}_\bml$: as we take steps in $\bth$-space we can simultaneously
take steps in $\bml$-space, thus adaptively choosing the form the ${\theta_\alpha}$ gradient estimator takes. 
This can be understood as an instance of an adaptive control variate \cite{ross};
consider the solution ${\bm v}^{\theta_\alpha}_{\bml}$ as well as a fixed reference solution ${\bm v}^{\theta_\alpha}_{0}$:
\begin{equation}
\label{eqn:avfmaster}
\begin{split}
\nabla_{\theta_\alpha} \mathcal{L} &= \E_{q_\bth(\bz)} \left[ \nabla_\bz f \cdot {\bm v}^{\theta_\alpha}_{\bml} \right] \\
&=
\E_{q_\bth(\bz)} \left[ \nabla_\bz f \cdot {\bm v}^{\theta_\alpha}_{0} \right] + 
\E_{q_\bth(\bz)} \left[ \nabla_\bz f \cdot \left({\bm v}^{\theta_\alpha}_{\bml} -  {\bm v}^{\theta_\alpha}_{0}\right) \right] 
\end{split}
\end{equation}
The final expectation is identically equal to zero so the integrand is a control variate. If we
choose $\bml$ appropriately we can reduce the variance of our gradient estimator.
In order to specify a complete algorithm we need to answer two questions:
i) how should we adapt $\bml$?; and ii) what is an appropriate family of solutions ${\bm v}^{\theta_\alpha}_{\bml}$?
We now address each of these in turn.

\subsection{Adapting velocity fields}

\newcommand\reallywidehat[1]{%
\savestack{\tmpbox}{\stretchto{%
  \scaleto{%
    \scalerel*[\widthof{\ensuremath{#1}}]{\kern-.6pt\bigwedge\kern-.6pt}%
    {\rule[-\textheight/2]{1ex}{\textheight}}
  }{\textheight}%
}{0.5ex}}%
\stackon[1pt]{#1}{\tmpbox}%
}

Whenever we compute a gradient estimate via Eqn.~\ref{eqn:estimator} we can use the same $N_s$
sample(s) $\bz_i \sim q_\bth$ to form an estimate of the variance of the gradient estimator:
\begin{equation}
\label{eqn:totgradvar}
\begin{split}
&{\rm Var}\!\left(\frac{\reallywidehat{\partial \mathcal{L}}}{\partial {\bth}}\right)  \equiv
\sum_{\alpha} {\rm Var}\!\left(\frac{\reallywidehat{\partial \mathcal{L}}}{\partial {\theta_\alpha}}\right) = \\ \nonumber
&\tfrac{1}{N_s}\sum_{\alpha} \sum_{\bz_i} \left[ (\nabla_\bz f \cdot {\bm v}_\bml^{\theta_\alpha})^2  \right]
 - \sum_{\alpha}\left( \tfrac{1}{N_s} \sum_{\bz_i} \left[ \nabla_\bz f \cdot {\bm v}_\bml^{\theta_\alpha}  \right] \right )^2 \nonumber
 \end{split}
 \end{equation} 
In direct analogy to the adaptation procedure used in (for example) \cite{ruiz2016overdispersed} we can adapt $\bml$ by following
noisy gradient estimates of this variance:
\begin{equation}
\label{eqn:lambdagrad}
\begin{split}
\nabla_\bml {\rm Var}\!\left(\frac{\reallywidehat{\partial \mathcal{L}}}{\partial {\bth}}\right) = \tfrac{1}{N_s}  \sum_\alpha
\sum_{\bz_i}\left[ \nabla_\bml (\nabla_\bz f \cdot {\bm v}_\bml^{\theta_\alpha})^2  \right]
\end{split}
\end{equation}
In this way $\bml$ can be adapted to reduce the gradient variance during the course of optimization.
See Algorithm \ref{avfalgobox} for a summary of the complete algorithm in the case where single-sample
gradient estimates are used at each iteration.

\begin{algorithm}
 {\bf Initialize} $\bth$, $\bml$, and choose step sizes $\epsilon_\bth, \epsilon_\bml$. \\
 \For{$i=1,2,...,N_{\rm{steps}}$}{
  1. Sample $\bz_i \sim q_\bth$\\
  2. Compute $g(\bth)^{\theta_\alpha}_i \equiv \nabla_\bz f(\bz_i) \cdot {\bm v}^{\theta_\alpha}_{\bml} \;\; {\rm for\; all\;} \alpha$ (cf.~Eqn.~\ref{eqn:avfmaster}) \\
  3. Compute $\bm{g}(\bml)_i \equiv \nabla_\bml {\rm Var}\!\left(\frac{\partial \mathcal{L}}{\partial {\bth}}\right)$ (cf.~Eqn.~\ref{eqn:lambdagrad}) \\
  4. Take a (noisy) gradient step $\bth \rightarrow \bth + \epsilon_\bth\bm{g}(\bth)_i$ \\
  5. Take a (noisy) gradient step $\bml \rightarrow \bml + \epsilon_\bml \bm{g}(\bml)_i$
 }
 \caption{Stochastic optimization with Adaptive Velocity Fields. Note that steps 2 and 4 occur implicitly for the reparameterization trick
 estimator. What sets Adaptive Velocity Fields apart is steps 3 and 5.}
 \label{avfalgobox}
\end{algorithm}

\subsection{Parameterizing velocity fields}

Assuming we have obtained (at least) one solution ${\bm v}^\theta_0$ to the transport equation Eqn.~\ref{eqn:transport}, 
parameterizing a family of solutions ${\bm v}^\theta_{\bml}$ is in principle easy. 
We simply solve the null equation, $ \nabla_\bz \cdot \left( q_\bth \tilde{{\bm v}}^\theta \right)=0$,
which is obtained from the transport equation by dropping the source term $\frac{\partial}{\partial \theta} q_\bth$.
Then for a given family of solutions to the null equation, $\tilde{{\bm v}}^\theta_{\bml}$, we get a solution 
${\bm v}^\theta_{\bml}$ to the transport equation by simple 
addition, i.e.~${\bm v}^\theta_{\bml} = {\bm v}^\theta_0 + \tilde{{\bm v}}^\theta_{\bml}$.
At first glance solving the null equation appears to be easy, since \emph{any} divergence-free vector field immediately yields a solution:
\vspace{-1mm}
\begin{equation}
\nonumber
\nabla_\bz \cdot \bm{w}_{\bml}=0 \;\; {\rm and} \;\; \tilde{{\bm v}}^\theta_{\bml} = \frac{\bm{w}_{\bml}}{q_\bth} \;\; \Rightarrow \;\; \nabla_\bz \cdot \left( q_\bth \tilde{{\bm v}}^\theta_{\bml} \right)=0
\end{equation}
However, in order to construct general purpose gradient estimators we need to impose appropriate boundary conditions on the velocity field;
for more on this subtlety we refer the reader to the supplementary materials.

In order to demonstrate how to construct suitable null solutions we henceforth focus on a particular family of distributions $q_\bth(\bz)$,
namely  \emph{elliptical} distributions. Elliptical distributions are probability distributions whose 
density is of the form $q(\bz) \propto g(\bz^{\rm T} \bm{\Sigma}^{-1} \bz)$ for some scalar density $g(\cdot)$ and positive definite
symmetric matrix $\bm{\Sigma}$. They include, for example,
the multivariate Normal distribution and the multivariate t-distribution. For concreteness, in the following we focus exclusively on the 
multivariate Normal distribution. We stress, however, that the resulting adaptive gradient estimators are applicable to any elliptical
distribution. We refer to the supplementary materials for a description of the specific case of the multivariate t-distribution.

\subsection{Null Solutions for the Multivariate Normal Distribution} 

Consider the multivariate Normal distribution in $D$ dimensions parameterized via a mean $\bmu$ and Cholesky factor $\bm{L}$. We would like
to compute derivatives w.r.t.~$L_{ab}$. While the reparameterization trick is applicable in this case, we can potentially get lower variance
gradients by using Adaptive Velocity Fields.
As we show in the supplementary materials, a simple solution to the corresponding null transport equation is given by
\begin{equation}
\label{eqn:nullsolnmvn}
\bm{\tilde{v}}^{L_{ab}}_{\bm{A}} = \bm{L}\bm{A}^{ab}\bm{L}^{-1} (\bz -\bmu)
\end{equation}
where $\bml \equiv \{ \bm{A}^{ab} \}$ is an arbitrary collection of antisymmetric $D \times D$ matrices (one for each $a,b$).
This is just a linear vector field and so it is easy to manipulate
and (relatively) cheap to compute. Geometrically, for each entry of $\bm{L}$, this null solution corresponds to an infinitesimal rotation in whitened coordinates $\bm{\tilde{z}} = \bm{L}^{-1} (\bz-\bmu)$. Note that $\bm{\tilde{v}}^{L_{ab}}_{\bm{A}} $ is \emph{added} to the reference solution $\bm{v}_0^{L_{ab}}$, and
this is \emph{not} equivalent to rotating $\bm{v}_0^{L_{ab}}$. 

Since $a,b$ runs over $\frac{D(D+1)}{2}$ pairs of indices and each $\bm{A}^{ab}$ has $\frac{D(D-1)}{2}$ free parameters,
this space of solutions is quite large. Since we will be adapting $\bm{A}^{ab}$ via noisy gradient estimates---and because
we would like to limit the computational cost---we choose to parameterize a smaller subspace of solutions. Empirically
we find that the following parameterization works well:
\begin{equation}
A_{jk}^{ab} = \sum_{\ell=1}^M B_{\ell a} C_{\ell b} \left(\delta_{aj}\delta_{bk} - \delta_{ak}\delta_{bj}\right)
\end{equation}
Here each $M \times D$ matrix $\bm{B}$ and $\bm{C}$ is arbitrary and the hyperparameter $M$
 allows us to trade off the flexibility of our family of null solutions with computational cost (typically $M \ll D$).
 
 The computational complexity of using Adaptive Velocity Field gradients with this class of parameterized velocity fields 
 (including the $\bm{A}$ update equations) 
is $\mathcal{O}(D^3 + MD^2)$ per gradient step.
This should be compared to the $\mathcal{O}(D^2)$ cost of the reparameterization trick gradient and the $\mathcal{O}(D^3)$
cost of the OMT gradient (see Sec.~\ref{sec:related} for a discussion of the latter).\footnote{Note, however, that the computational 
complexity of the AVF gradient is somewhat misleading in that the 
$\mathcal{O}(D^3)$ term arises from matrix
multiplications, which tend to be quite fast. By contrast the OMT gradient estimator involves a singular value decomposition, which
tends to be substantially more expensive than a matrix multiplication on modern hardware. }
 We explore the performance of the resulting Adaptive Velocity Field (AVF) gradient estimator in Sec.~\ref{sec:avfexp}.


\section{Pathwise Derivatives for Multivariate Mixture Distributions}
\label{sec:mix}

In this section we use the transport equation Eqn.~\ref{eqn:transport} to construct pathwise gradient
estimators for mixture distributions. Note that this section is logically independent from the previous section, 
although---as we discuss briefly in the supplementary materials---Adaptive Velocity Fields can be readily combined with the gradient estimators discussed here.

Consider a mixture of $K$ multivariate distributions in $D$ dimensions:
\vspace{-3mm}
\begin{equation}
q_\bth(\bz) = \sum_{j=1}^K \pi_j q_{\bth_{j}}(\bz)
\end{equation}
In order to compute pathwise derivatives of $\bz \sim q_\bth(\bz)$ we need to solve
the transport equation w.r.t.~the mixture weights $\pi_j$ as well as the component parameters $\bth_{j}$. 
For the derivatives w.r.t.~$\bth_j$ we can just repurpose the velocity fields of the individual components. That is,
if $\bm{v}^{\bth_j}_{\rm{single}}$ is a solution of the single-component transport equation 
$\big($i.e.~$\frac{\partial q_{\bth_j}}{\partial \bth_j} + \nabla \cdot(q_{\bth_j} \bm{v}^{\bth_j}_{\rm{single}}) = 0\big)$, 
%
%
then 
\begin{equation}
\label{eqn:multirepurpose}
\bm{v}^{\bth_j} = \frac{ \pi_j q_{\bth_j}} {q_\bth} \bm{v}^{\bth_j}_{\rm{single}}
\end{equation}
is a solution of the multi-component transport equation.\footnote{We note that the form of Eqn.~\ref{eqn:multirepurpose} implies that we can easily introduce Adaptive Velocity Fields for each individual component (although the cost of doing so may be prohibitively high for a large number of components).} See the supplementary materials for the (short) derivation.
For example we can readily compute Eqn.~\ref{eqn:multirepurpose} if each component distribution is reparameterizable. 
Note that a typical gradient estimator for, say, a mixture of Normal distributions first samples the discrete component 
$k \in [1, ..., K]$ and then samples $\bz$ conditioned on $k$. This results in a pathwise derivative for $\bz$ 
that only depends on $\bmu_k$ and $\bm{\sigma}_k$. By contrast the  gradient computed via Eqn.~\ref{eqn:multirepurpose}
will result in a gradient for all $K$ component parameters for every sample $\bz$. We explore this difference experimentally in Sec.~\ref{sec:baseball}.

\subsection{Mixture weight derivatives}

Unfortunately, solving the transport equation for the mixture weights is in general much more difficult, since
the desired velocity fields need to coordinate mass transport among all $K$ components.\footnote{For example,
we expect this to be particularly difficult if some of the component distributions belong to different families of 
distributions, e.g. a mixture of Normal distributions and Wishart distributions.} 
We now describe a formal construction for solving the transport equation for the mixture weights.
For each pair of component distributions $j$, $k$ consider the transport equation 
\begin{equation}
\label{eqn:pairwisetransport}
q_{\bth_j} - q_{\bth_k}+  \nabla_\bz \cdot \left( q_\bth \tilde{\bm{v}}^{jk} \right)=0
\end{equation}
Intuitively, the velocity field $\tilde{\bm{v}}^{jk}$ moves mass from component $k$ to component $j$. 
We can superimpose these  solutions to form
\vspace{-1mm}
\begin{equation}
\label{eqn:vlogits}
\bm{v}^{\ell_j} = \pi_j \sum_{k\ne j} \pi_k \tilde{\bm{v}}^{jk}
\end{equation}
As we show in the supplementary materials, the velocity field $\bm{v}^{\ell_j}$ yields an estimator for the 
derivative w.r.t.~the softmax logit $\ell_j$ that corresponds to component $j$.\footnote{That is with respect to $\ell_j$ where
$\pi_j = e^{\ell_j} / \sum_k e^{\ell_k}$.} Thus provided we can find solutions $\tilde{\bm{v}}^{jk}$ to Eqn.~\ref{eqn:pairwisetransport}
for each pair of components $j$, $k$ we can form a pathwise gradient estimator for the mixture weights.
We now consider one particular situation where this recipe can be carried out. We leave a discussion of other solutions---including
a solution for a mixture of Normal distributions with arbitrary diagonal covariances---to the supplementary materials.

\subsection{Mixture of Multivariate Normals with Shared Diagonal Covariance}
\label{sec:sharedcov}

Here each component distribution is specified by $q_{\bth_j}(\bz) = \mathcal{N}(\bz | \bmu_j, \bm{\sigma})$, i.e.~each
component distribution has its own mean vector $\bmu_j$ but all $K$ components share
the same (diagonal) square root covariance $\bm{\sigma}$.\footnote{To keep the equations compact we consider a diagonal covariance matrix. The
resulting velocity field is readily transformed to a rotated coordinate system in which the (single shared) covariance matrix is arbitrary.}
We find that a solution to Eqn.~\ref{eqn:pairwisetransport} is given by
\begin{equation}
\label{eqn:arbmumixsoln}
 \tilde{\bm{v}}^{jk} = \frac{\left(\Phi(\tilde{z}_\parallel^{jk} \!-\! \tilde{\mu}^{jk}_\parallel ) - 
                                                  \Phi(\tilde{z}_\parallel^{jk} \!+\! \tilde{\mu}^{kj}_\parallel ) \right)
 \phi(||\tilde{\bz}_\bot^{jk}||^2)}{(2\pi)^{D/2-1}q_\bth \prod_{i=1}^D \sigma_i} \bm{\sigma}  \hat{\bmu}^{jk} 
 \end{equation}
where $\bm{\sigma}  \hat{\bmu}^{jk}$ indicates matrix multiplication by ${\rm diag}(\bm{\sigma})$, $\tilde{\bz} \equiv \bz \odot \bm{\sigma}^{-1}$,
and where the quantities
$\{ \tilde{\bmu}^j, \hat{\bmu}^{jk}, \tilde{\mu}^{jk}_\parallel, \tilde{z}_\parallel^{jk}, \tilde{\bz}_\bot^{jk} \}$
are defined in the supplementary materials.
Here $\Phi(\cdot)$ is the CDF of the unit Normal distribution and $\phi(\cdot)$ is the corresponding probability density.
Geometrically, the velocity field in Eqn.~\ref{eqn:arbmumixsoln} moves mass along lines parallel to $\bmu_j - \bmu_k$.
Note that the gradient estimator resulting from Eqn.~\ref{eqn:arbmumixsoln} has cost $\mathcal{O}(DK^2)$.\footnote{As detailed in the supplementary materials, the pathwise gradient estimators for the familes of mixture distributions we consider all have cost $\mathcal{O}(DK)$ or $\mathcal{O}(DK^2)$.} We explore the performance of the resulting pathwise gradient estimators in Sec.~\ref{sec:mixexp}.


\section{Related Work}
\label{sec:related}

There is a large body of work on constructing
gradient estimators with reduced variance, much of which can be understood in terms of control variates \cite{ross}:
for example, \cite{mnih2014neural} constructs neural baselines for score function gradients; \cite{schulman2015gradient}
discuss gradient estimators for stochastic computation graphs and their Rao-Blackwellization;
 and \cite{tucker2017rebar, grathwohl2017backpropagation}
construct adaptive control variates for discrete random variables. Another example of this line of work is \cite{miller2017reducing},
where the authors construct control variates that are applicable when $q_\bth(z)$ is a \emph{diagonal} Normal distribution 
and that rely on Taylor expansions of the test function $f(\bz)$.\footnote{Also, in their approach 
variance reduction for scale parameter gradients $\nabla_{\bm{\sigma}}$ necessitates a multi-sample estimator (at least for high-dimensional models where computing the diagonal of the Hessian is expensive).}
In contrast our AVF gradient estimator for the multivariate Normal has about half the computational cost, since it does not rely on
second order derivatives.
Another line of work constructs partially reparameterized gradient estimators for cases where the reparameterization
trick is difficult to apply \cite{ruiz2016generalized, naesseth2017reparameterization}. 
In \cite{graves2016stochastic}, the author constructs pathwise gradient estimators for mixtures of diagonal Normal distributions.
Unfortunately, the resulting estimator is expensive, relying on a recursive computation that scales with the dimension of the sample space.
Another approach to mixture distributions is taken in \cite{dillon2018quadrature}, where the authors use quadrature-like constructions to form continuous distributions that are reparameterizable.
Ref.~\cite{roeder2017sticking} considers a gradient estimator for mixture distributions in which all $K$ components are summed out. 
Our work is closest to \cite{beyond}, which uses
the transport equation to derive an `OMT' gradient estimator for the multivariate Normal distribution whose velocity field is optimal in the
sense of optimal transport. This gradient estimator can exhibit low variance but has $\mathcal{O}(D^3)$ computational complexity due to an (expensive)
singular value decomposition. In addition the OMT gradient estimator is \emph{not} adapted to the test function at hand; rather it has one fixed form. 

As this manuscript was being completed, we became aware of reference \cite{figurnov2018implicit}, which has some overlap with this
work. In particular, \cite{figurnov2018implicit} describes an interesting recipe for computing pathwise gradients that can be applied to
mixture distributions. A nice feature of this recipe (also true of \cite{graves2016stochastic}), is that it can be applied to mixtures of Normal distributions with arbitrary covariances.
A disadvantage is that it can be expensive, exhibiting a $\mathcal{O}(D^2)$ computational cost even for a mixture of 
Normal distributions with diagonal covariances. In contrast the gradient estimators constructed in Sec.~\ref{sec:mix} by solving
the transport equation are linear in $D$. We provide a comparison to their approach in Sec.~\ref{sec:synthmix}.


\section{Experiments}
\label{sec:exp}

\begin{figure}[t!]
    \centering
    \begin{subfigure}{.49\textwidth}
        \includegraphics[width=.9\textwidth]{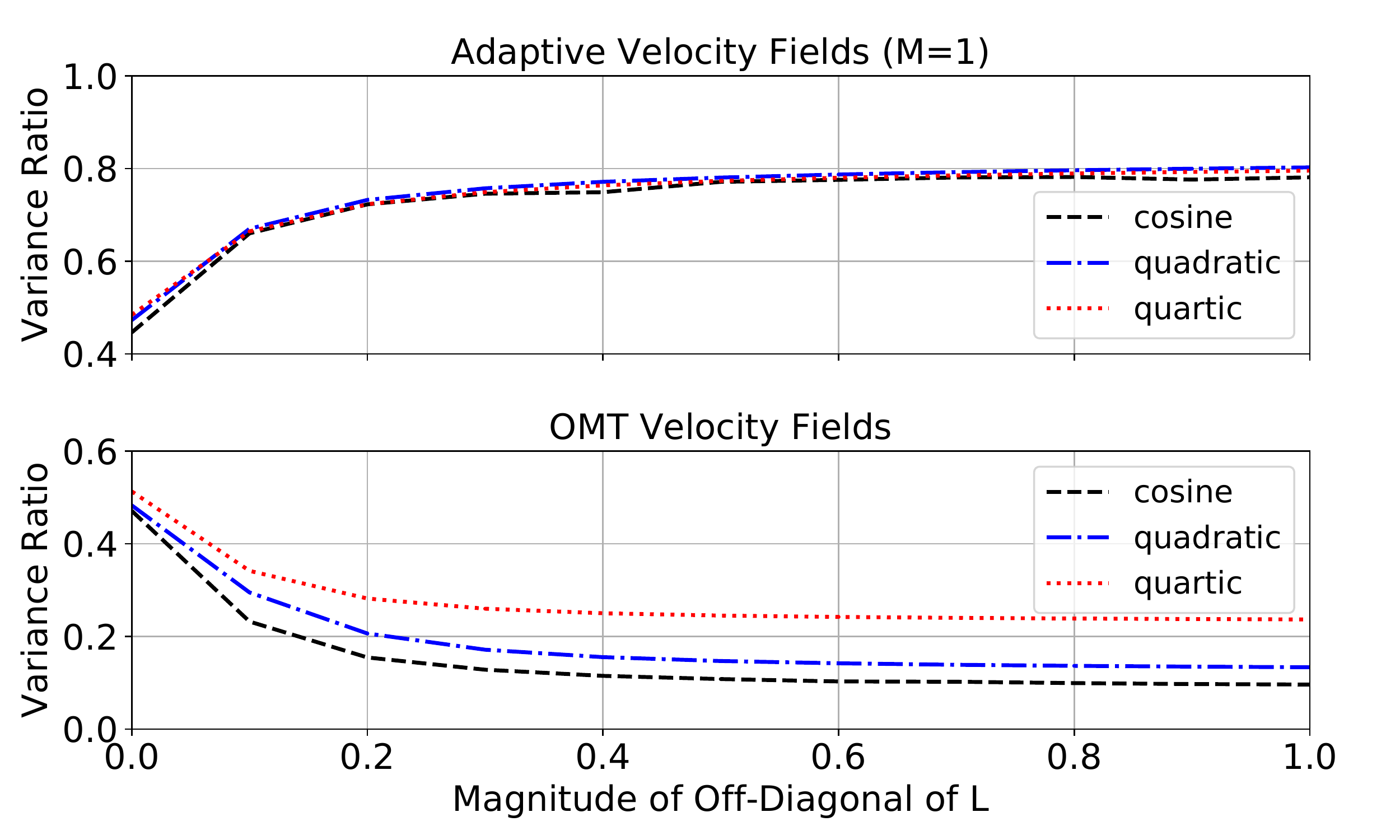}
        \caption{We compare the OMT and AVF gradient estimators for the multivariate Normal distribution to the RT estimator for three test functions.
        The horizontal axis controls the magnitude of the off-diagonal elements of the Cholesky factor $\bm{L}$.
The vertical axis depicts the ratio of the mean variance of the given estimator to that of the RT estimator for the off-diagonal elements of $\bm{L}$.}
\label{fig:mvnreach}
    \end{subfigure}\hfill
    \begin{subfigure}{.49\textwidth}
        \includegraphics[width=.95\textwidth]{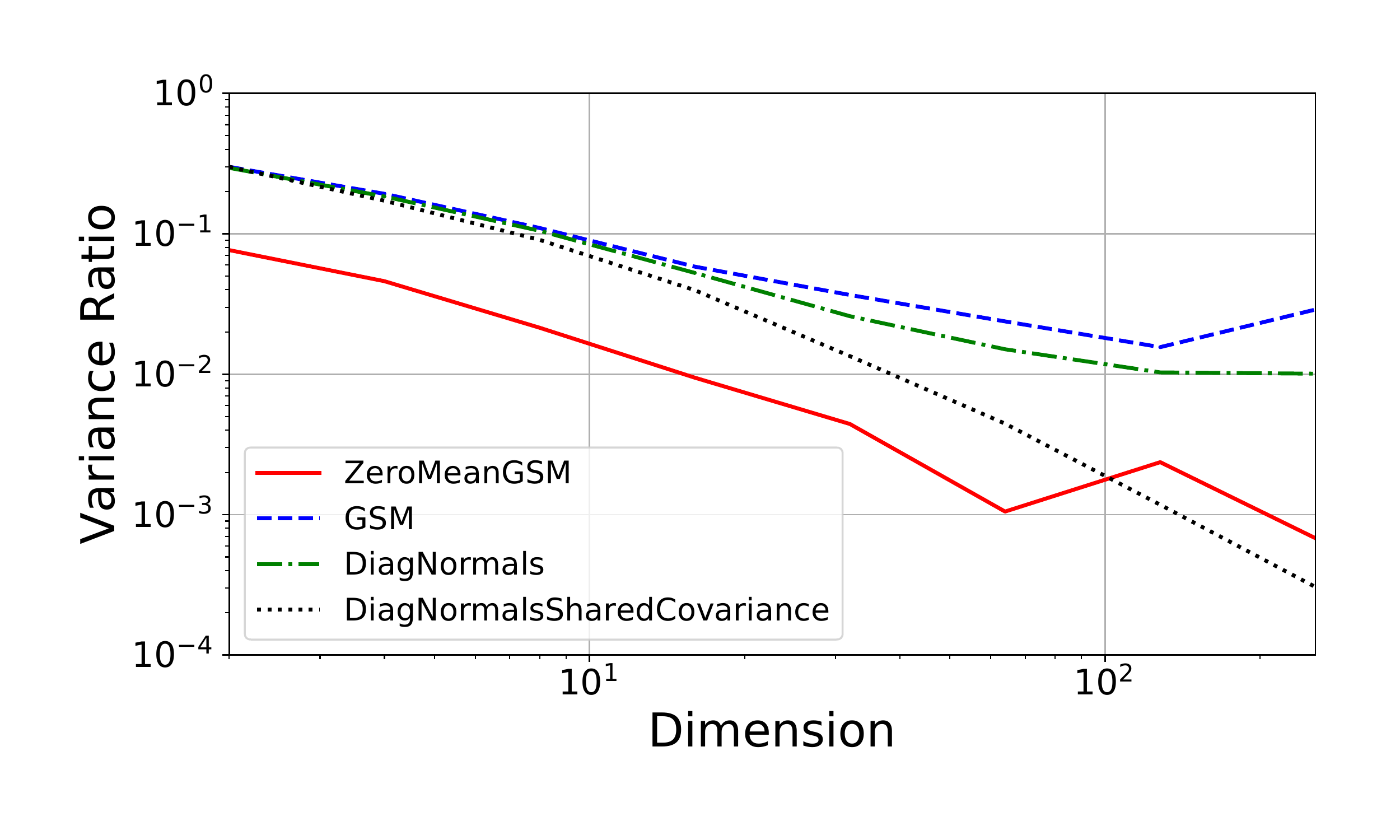}
\caption{We compare the variance of pathwise gradient estimators for various mixture distributions with $K=10$
 to the corresponding score function estimator.
The vertical axis depicts the ratio of the logit gradient variance of the pathwise estimator to that of the score function estimator for the test function
$f(\bz)=||\bz||^2$.}
\label{fig:mixreach}
    \end{subfigure}
        \begin{subfigure}{.49\textwidth}
        \includegraphics[width=.95\textwidth]{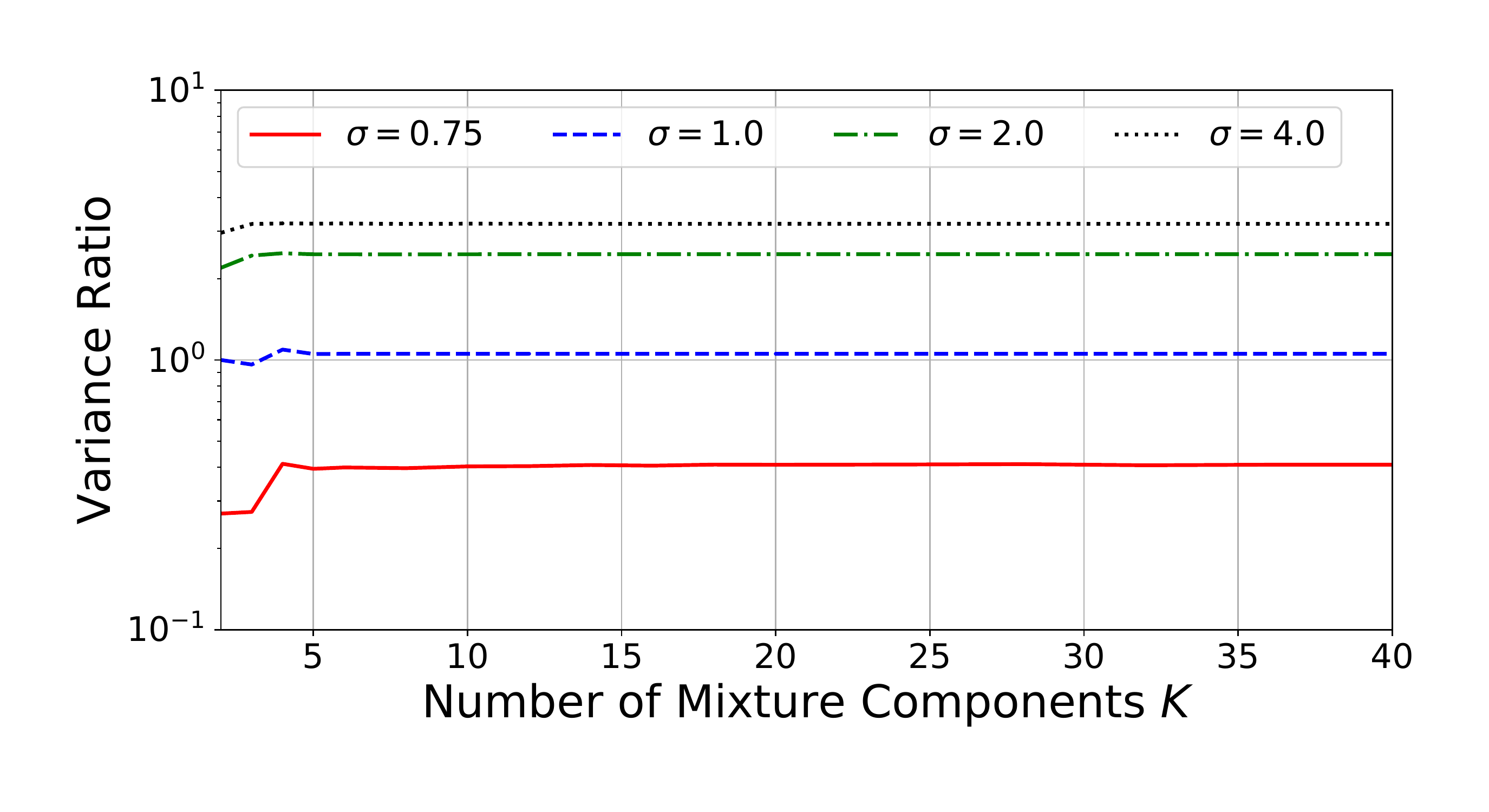}
\caption{We compare the variance of our pathwise gradient estimator for mixtures of diagonal Normal distributions to the one described in ref.~\cite{figurnov2018implicit}. Here $D=2$, $\sigma$ controls the covariance of each component distribution, and the vertical axis is the ratio of variances (with our estimator in the denominator). See supplementary materials for details.}
\label{fig:mixreach2}
    \end{subfigure}
    \caption{We compare the gradient variances for different gradient estimators and synthetic test functions. See Sec.~\ref{sec:synthavf} and Sec.~\ref{sec:synthmix} for details.}
\end{figure}

We conduct a series of experiments using synthetic and real world data to validate the performance of the gradient
estimators introduced in Sec.~\ref{sec:avf} \& \ref{sec:mix}. Throughout we use single sample estimators.
See Sec.~\ref{sec:expsupp} in the supplementary materials for details on experimental setups. 
Open source implementations of all our gradient estimators are available at \texttt{https://git.io/fhbqH}.  All variational inference experiments were implemented using the Pyro probabilistic programming language \cite{bingham2018pyro}.


\subsection{Adaptive Velocity Fields}
\label{sec:avfexp}

\subsubsection{Synthetic test functions}
\label{sec:synthavf}

In Fig.~\ref{fig:mvnreach} we use synthetic test functions to illustrate the variance reduction that is achieved
with AVF gradients for the multivariate Normal distribution as compared to the reparameterization trick and
OMT gradients from \cite{beyond}. The dimension is $D=50$; the results are qualitatively similar for different dimensions.
Note that, as demonstrated here, whether AVF outperforms OMT is problem dependent.

\subsubsection{Gaussian Process Regression}
\label{sec:co2gp}

\begin{figure}[t]
\begin{center}
\centerline{\includegraphics[width=.95\columnwidth]{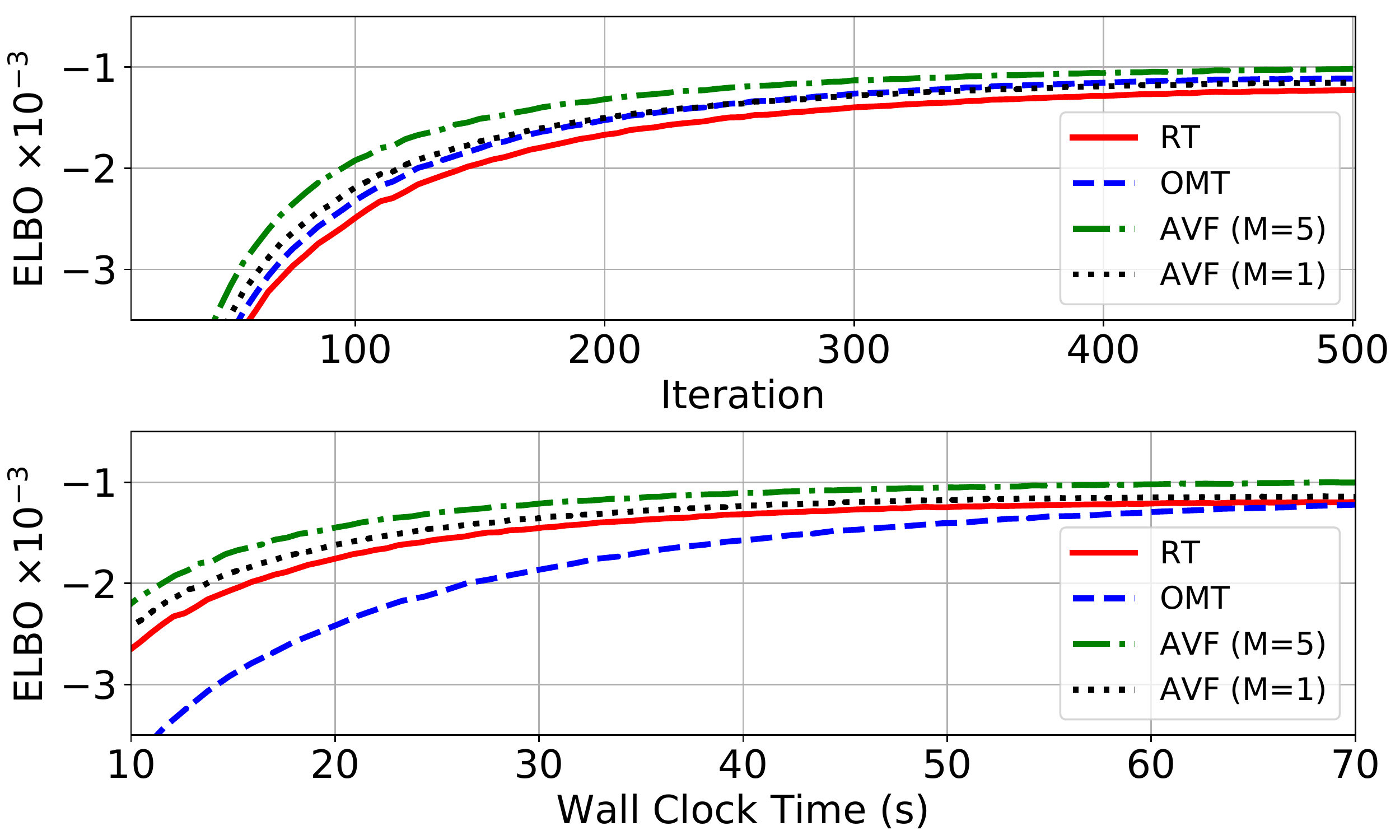}}
\caption{We compare the performance of various pathwise gradient estimators on the GP regression task in Sec~\ref{sec:co2gp}. The AVF
gradient estimator with $M=5$ is the clear winner both in terms of wall clock time and the final ELBO achieved.}
\label{fig:co2gp}
\end{center}
\end{figure}

We investigate the performance of AVF gradients for the multivariate Normal
distribution in the context of a Gaussian Process regression task reproduced from \cite{beyond}. 
We model the Mauna Loa ${\rm CO}_2$ 
data from \cite{keeling2004atmospheric} considered
in \cite{rasmussen2004gaussian}. 
We fit the GP using a single-sample Monte Carlo ELBO gradient estimator and all $N=468$ data points (so $D=468$).
Both the OMT and AVF gradient estimators achieve higher ELBOs than the RT estimator (see Fig.~\ref{fig:co2gp}); 
however, the OMT estimator does so at substantially increased computational cost ($\sim$1.9x), while the AVF estimator
for $M=1$ ($M=5$) requires only $\sim$6\% ($\sim$11\%) more time per iteration. By iteration 250 the AVF gradient estimator 
with $M=5$ has attained the same ELBO that the RT estimator attains at iteration 500.

\vspace{-3pt}


\subsection{Mixture Distributions}
\label{sec:mixexp}

\subsubsection{Synthetic test function}
\label{sec:synthmix}

In Fig.~\ref{fig:mixreach} we depict the variance reduction of pathwise gradient estimators for various mixture distributions compared to the corresponding score function estimator. We find that the typical variance reduction has magnitude $\mathcal{O}(D)$. The results are qualitatively similar for different
numbers of components $K$. See the supplementary materials for details on the different families of mixture distributions.
In Fig.~\ref{fig:mixreach2} we compare the gradient estimator for mixtures of Normal distributions described in Sec.~\ref{sec:sharedcov} to the estimator described in \cite{figurnov2018implicit}. We find that our estimator exhibits better performance when the mixture components have more overlap ($\sigma \gg 0$), while their estimator exhibits better performance in the opposite regime ($\sigma \to 0$). Since their estimator has a cost that is quadratic in the dimension $D$, while our estimator has a cost that is linear in the dimension $D$, which estimator is to be preferred will be problem specific.

\subsubsection{Baseball Experiment}
\label{sec:baseball}

\begin{figure}[t]
    \centering
    \begin{subfigure}{0.49\textwidth}
        \includegraphics[width=.99\textwidth]{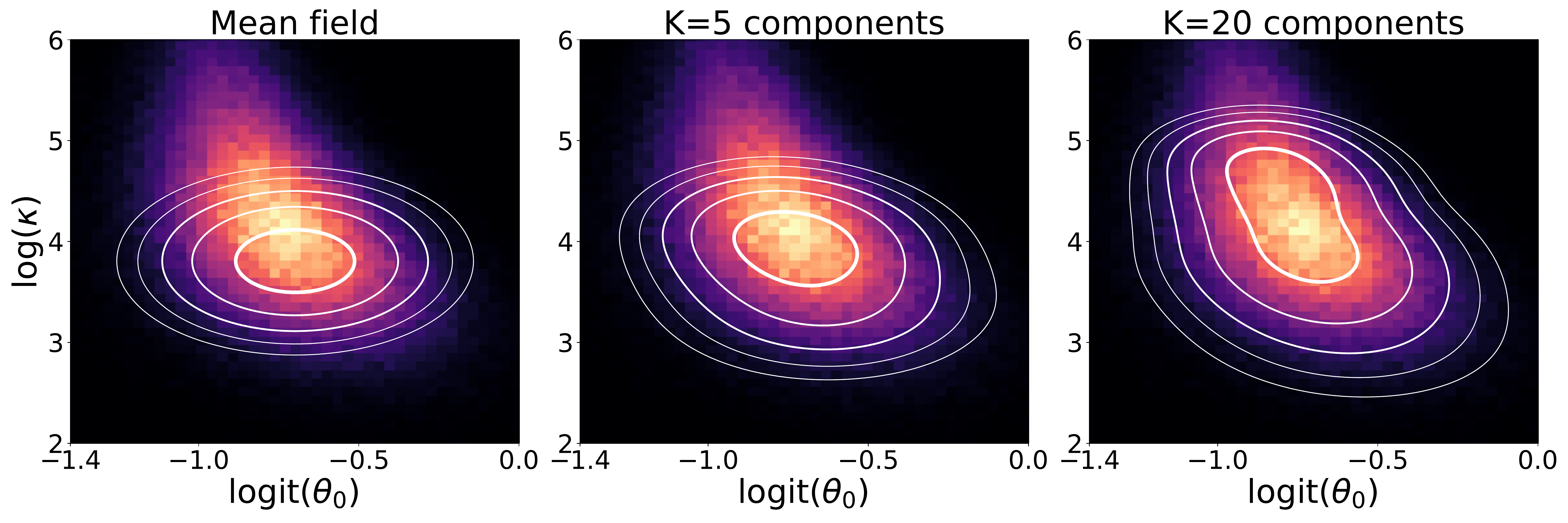}
        \caption{Two-dimensional cross-sections of three approximate posteriors; each cross-section includes one global latent variable, $\log( \kappa)$, and one local latent variable, ${\rm logit}( \theta_0)$. The white density contours correspond to the different variational approximations, and the background depicts the approximate posterior computed with HMC.  The quality of the variational approximation improves as we add more mixture components.}
\label{fig:bivariate}
    \end{subfigure}\hfill
    \begin{subfigure}{0.49\textwidth}
        \includegraphics[width=.95\textwidth]{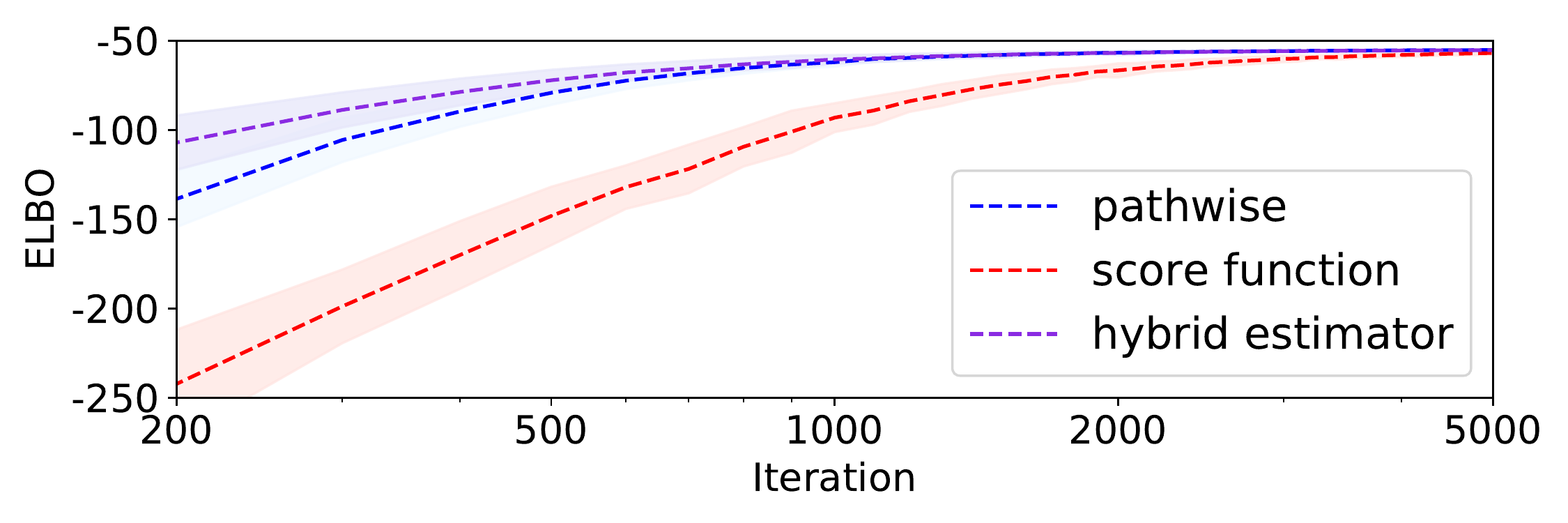}
\caption{ELBO training curves for three gradient estimators for $K=20$. The lower variance of the pathwise and hybrid gradient estimators speeds up learning. The dotted lines denote mean ELBOs over 50 runs and the bands indicate $1{\hbox{-}}\sigma$ standard deviations. 
Note that the figure would be qualitatively similar if the ELBO were plotted against wall clock time, since the estimators have comparable computational cost (within 10\%). }
\label{fig:baseball_elbo}
    \end{subfigure}
    \caption{Variational approximations and ELBO training curves for the model in Sec.~\ref{sec:baseball}.}
\end{figure}

We consider a model for repeated binary trial data using the data in \cite{efron1975data} and the modeling setup in \cite{stanmanual} with partial pooling. 
The model has two global latent variables and 18 local latent variables so that the posterior is 20-dimensional. While mean field SGVI gives reasonable results for this model, it is not able to capture the detailed structure of the exact posterior as computed with the NUTS HMC implementation in Stan \cite{hoffman2014no, carpenter2017stan}. To go beyond mean field we consider variational distributions that are mixtures of $K$ diagonal Normal distributions.\footnote{Cf.~the experiment in \cite{miller2016variational}} Adding more components gives a better approximation to the exact posterior, see Fig.~\ref{fig:bivariate}. Furthermore, using pathwise derivatives in the ELBO gradient estimator leads to faster convergence, see Fig.~\ref{fig:baseball_elbo}. Here the hybrid estimator
uses score functions gradients for the mixture logits but implements Eqn.~\ref{eqn:multirepurpose} for gradients w.r.t.~the component parameters.
Since there is significant overlap among the mixture components, Eqn.~\ref{eqn:multirepurpose} leads to substantial variance reduction.

\subsubsection{Continuous state space model}
\label{sec:ssm}

\begin{figure}[t]
    \centering
    \begin{subfigure}{0.49\textwidth}
        \includegraphics[width=.9\textwidth]{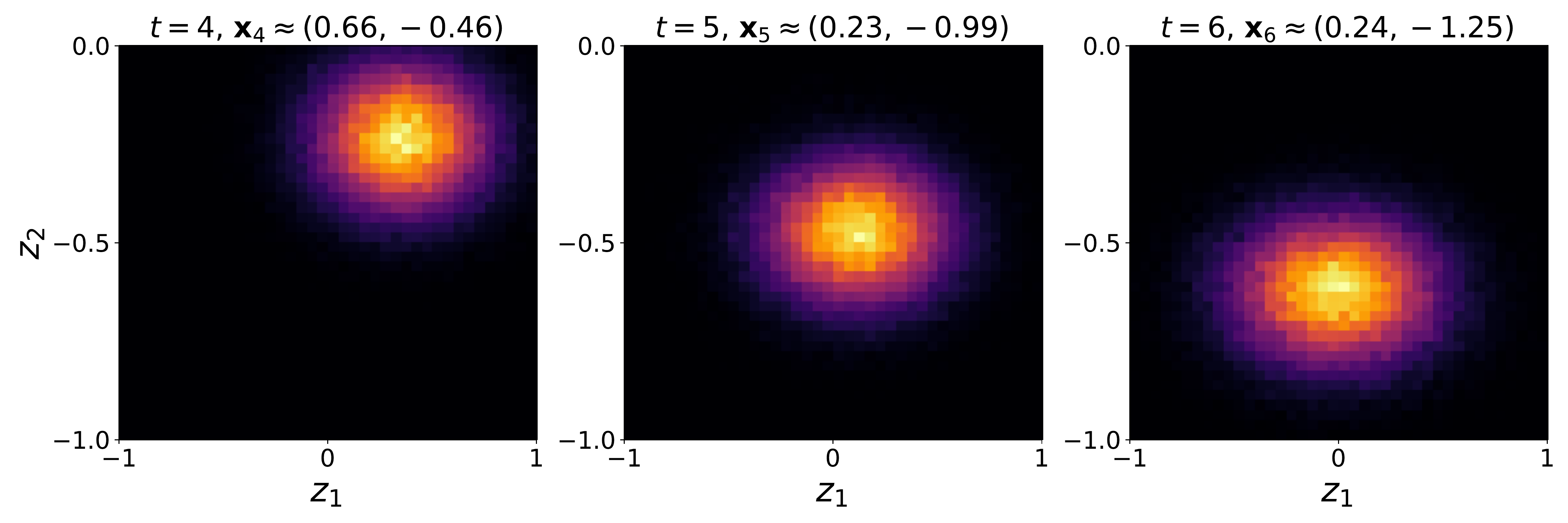}
        \caption{$K=1$ (no mixture)}
\label{fig:ssmnc1}
    \end{subfigure}\hfill
    \begin{subfigure}{0.49\textwidth}
        \includegraphics[width=.9\textwidth]{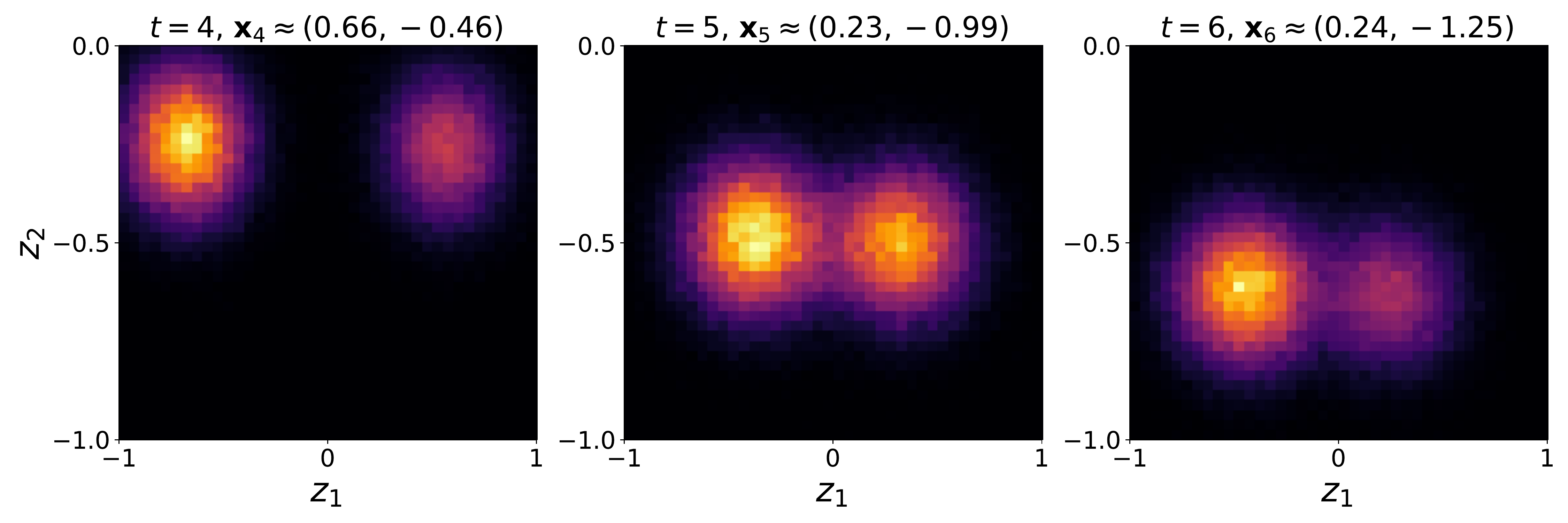}
                \caption{$K=2$ (mixture)}
\label{fig:ssmnc2}
    \end{subfigure}
\caption{Approximate marginal posteriors for the toy state space model in Sec.~\ref{sec:ssm} for a randomly chosen
test sequence for $t=4,5,6$. The mixture distributions that serve as components of the variational family on the bottom help capture the multi-modal structure of the posterior.}
   \label{fig:ssm}
\end{figure}

We consider a non-linear  state space model with observations $\bx_t$ and 
random variables $\bz_t$ at each time step (both two dimensional). 
Since the likelihood is of the form $p(\bx_t | \bz_t) = \mathcal{N}(\bx_t |\bm{\mu}(\bz_t), \sigma_x \mathbb{1}_2)$ with 
$\bm{\mu}(\bz_t) = (z_{1t}^2, 2z_{2t})$, the posterior over $\bz_{1:T}$ is highly multi-modal.
We use SGVI to approximate the posterior over $\bz_{1:T}$. To better capture the multi-modality
we use a variational family that is a product of two-component mixture distributions of the form $q(\bz_t | \bz_{t-1}, \bx_{t:T})$, one at each
time step.
As can be seen from Fig.~\ref{fig:ssm}, the variational family makes use of the mixture distributions at its disposal to model
the multi-modal structure of the posterior, something that a variational family with a unimodal $q(\bz_t | \cdot)$ struggles to do. 
In the next section we explore the use of mixture distributions in a much richer time series setup.

\subsubsection{Deep Markov Model}
\label{sec:dmm}

We consider a variant of the Deep Markov Model (DMM) setup introduced in \cite{krishnan2017structured} on a polyphonic
music dataset. 
We fix the form of the model and vary the variational family and gradient estimator used. We consider a
variational family which includes a mixture distribution at each time step. In particular
each factor $q(\bz_t | \bz_{t-1}, \bx_{t:T})$ in the variational distribution is a mixture of $K$ diagonal Normals.
We compare the performance of the pathwise gradient estimator introduced in Sec.~\ref{sec:mix} to two variants of the Gumbel Softmax gradient estimator \cite{jang2016categorical,maddison2016concrete}, see Table \ref{table:dmm}. 
For a description of these two gradient estimators, which use the Gumbel Softmax trick to deal with mixture weight gradients,
see Sec.~\ref{sec:expsupp} in the supplementary materials.
Note that both Gumbel Softmax estimators are biased,
while the pathwise estimator is unbiased. 
We find that the Gumbel Softmax estimators---which induce $\mathcal{O}(2-8)$ times higher gradient variance for the mixture logits as compared to the pathwise estimator---are unable to achieve competitive 
test ELBOs.\footnote{We also found the Gumbel Softmax estimators to be less numerically stable, which prevented us from using more aggressive KL annealing schedules. This may have contributed to the performance gap  in Table  \ref{table:dmm}.} 

Note that while the numerical differences in Table \ref{table:dmm} are not  large ($\sim 0.05$ nats), 
they are comparable to the quantitative improvement that one can achieve in this context when equipping the variational family with normalizing flows \cite{tabak2013family, rezende2015variational}; for this particular task there appears to be little headroom above what can be attained with a (unimodal) Normal variational family.
\begin{table}[h!]
\begin{center}
    \begin{tabu}{|c|[1pt]c|c|c|}    \hline
   \cellcolor[gray]{0.75} & \multicolumn{3}{c|}{\small Gradient Estimator \cellcolor[gray]{0.95}} \\  \hline
    \small Variational Family \cellcolor[gray]{0.95} & \small Pathwise & \small GS-Soft  & \small GS-Hard \\  \tabucline[1pt]{-}
    \small K=2 &\small  \bf{-6.84} & \small -6.89 & \small -6.88 \\ \hline
    \small K=3 &\small  \bf{-6.82} & \small -6.87 & \small -6.85 \\ \hline
    \end{tabu}
\end{center}    
     \caption{Test ELBOs for the DMM in Sec.~\ref{sec:dmm}. Higher is better. 
     For comparison: we achieve a test ELBO of -6.90 for a variational
     family in which each $q(\bz_t | \cdot)$ is a diagonal Normal.}
     \label{table:dmm}
\end{table}

\subsubsection{VAE}
\label{sec:vae}

We conduct a simple experiment with a VAE \cite{kingma2013auto,rezende2014stochastic} trained on MNIST.
We fix the form of the model and vary the gradient estimator used.
The variational family is of the form $q(\bz | \bx)$, where $q(\bz | \cdot)$ is a mixture of $K \in [3,4,5]$ diagonal Normals.
We find that on this task our gradient estimator achieves similar performance to the two Gumbel Softmax estimators and the score function estimator, see Table \ref{table:vae}.  This is not too surprising, since VAE posteriors tend to be quite sharp for any given $\bx$ so that gradient variance is not the bottleneck for learning.

\begin{table}[h!]
\begin{center}
    \begin{tabu}{|c|[1pt]c|c|c|c|}    \hline
   \cellcolor[gray]{0.75} & \multicolumn{4}{c|}{ \small Gradient Estimator \cellcolor[gray]{0.95}} \\  \hline
     \small Var.~Family \cellcolor[gray]{0.95} &  \small Pathwise &  \small GS-Soft  &  \small GS-Hard &  \small SF \\  \tabucline[1pt]{-}
     \small K=3 & \small \bf{-95.88} &  \small -95.95 &  \small -95.96 &  \small  -95.98 \\ \hline
     \small  K=4 &  \small -95.89 &  \small \bf{-95.84} &  \small -95.88 &  \small -95.99 \\ \hline
     \small  K=5 &  \small -96.04 &  \small -95.94 &  \small \bf{-95.89} &  \small -95.98 \\ \hline
    \end{tabu}
\end{center}    
     \caption{Test ELBOs for the VAE in Sec.~\ref{sec:vae}. Higher is better.}
     \label{table:vae}
\end{table}


\section{Discussion}
\label{sec:discussion}

We have seen that the link between the transport equation and derivatives of expectations offers a powerful
tool for constructing pathwise gradient estimators. In the case of variational inference, 
we emphasize that---in addition to providing gradients with reduced 
variance---pathwise gradient estimators can enable quick iteration over complex models and variational families, since there
is no need to reason about ELBO gradients. The modeler need only construct a Monte Carlo estimator of the ELBO;
automatic differentiation will handle the rest.
Our hope is that an expanding toolkit of pathwise gradient estimators can contribute to 
the development of innovative applications of variational inference.

One limitation of our approach is that solving the transport equation for arbitrary families of distributions presents
a challenge. Of course, this limitation also applies to the reparameterization trick. While we acknowledge this limitation, we note that since: i) the Normal distribution is ubiquitous in machine learning; and ii) mixtures of Normals are universal approximators for continuous distributions, improving the gradient estimators that can be used in these two cases is already of considerable value. 

Given the richness of the transport equation and the great variety of multivariate distributions, there are several exciting possibilities
for future research. It would be of interest to identify more cases where we can construct null solutions to the transport equation, e.g.~generalizing Eqn.~\ref{eqn:nullsolnmvn} to higher orders. It could also be fruitful to consider other ways of
adapting velocity fields. In some cases this could lead to improved variance reduction, especially in the case of mixture
distributions, where the optimal transport setup in \cite{chen2017optimal} might provide a path to constructing alternative
gradient estimators. 

\bibliographystyle{abbrv} 
\bibliography{multipath}

\clearpage
\newpage
\appendix

\section{Supplementary Materials: Overview}

These supplementary materials are organized as follows. In Sec.~\ref{sec:pathest} we discuss general properties of pathwise gradient estimators
derived from the transport equation. In Sec.~\ref{sec:avfsupp} we give further details on Adaptive Velocity Fields. In Sec.~\ref{sec:mixsupp} we give further
details on our pathwise gradient estimators for mixture distributions, in particular describing velocity fields for four families of Normal mixtures.
Finally, in Sec.~\ref{sec:expsupp} we  describe the setup of the various experiments described in the main text.

\section{Pathwise Gradient Estimators and the Transport Equation}
\label{sec:pathest}

As discussed in the main text, a solution ${\bm v}^\theta$ to the transport equation allows us
to form an unbiased pathwise gradient estimator via
\begin{equation}
\label{eqn:estimator2}
\nabla_\theta \mathcal{L} = \E_{q_\bth(\bz)} \left[ \nabla_\bz f \cdot {\bm v}^\theta  \right]
\end{equation}
In order for this to be a sensible Monte Carlo estimator, we require that the variance is finite, i.e
\begin{equation}
\label{eqn:estimatorvar}
\mathbb{V}(\nabla_\theta \mathcal{L}) = \E_{q_\bth(\bz)} \left[ || \nabla_\bz f \cdot {\bm v}^\theta ||^2 \right] - ||\nabla_\theta \mathcal{L}||^2 < \infty
\end{equation}
In order for the derivation given in the main text to hold, we also require for ${\bm v}^\theta$ to be everywhere continuously differentiable
and that the surface integral $\oint_{S} (q_\bth f {\bm v}^\theta) \cdot \hat{\VF{n}} \dif S$ go to zero as $\dif S$ tends
towards the boundary at infinity. A natural way to ensure the latter condition for a large class of test functions is to require
the boundary condition
\begin{equation}
\label{eqn:bc}
q_\bth \bm{v}^{\pi_j} \to 0 \qquad {\rm as} \qquad ||\bz|| \to \infty \;\; ({\rm for\; all\; directions\;} \hat{\bm{z}})
\end{equation}
Note that this boundary condition is satisfied by all the gradient estimators proposed in this work.
Much of the difficulty in using the transport equation to construct pathwise gradient estimators is in
finding velocity fields that satisfy all these desiderata. 

For example consider a mixture of products of univariate distributions of the form:
\begin{equation}
\label{eqn:prodmix}
q_\bth(\bz) = \sum_{j=1}^K \pi_j q_{\bth_{j}}(\bz) \qquad {\rm with} \qquad q_{\bth_{j}}(\bz)= \prod_{i=1}^D q_{\bth_{ji}}(z_i)
\end{equation}
Here $j$ runs over the components and $i$ runs over the dimensions of $\bm{z}$. 
Note that a mixture of diagonal Normal distributions is a special case of Eqn.~\ref{eqn:prodmix}.
Suppose each $ q_{\bth_{ji}}$ has a CDF $F_{\bth_{ji}}$ that we have analytic control over.
Then we can form the velocity field
\begin{equation}
\label{eqn:badvel}
\bm{v}^{\pi_j}_i = - \frac{F_{\bth_{ji}} q_{j,-i}}{D q_\bth} \qquad {\rm with} \qquad q_{j,-i} \equiv \prod_{k\ne i} q_{\bth_{jk}}
\end{equation}
This is a solution to the transport equation for the mixture weight ${\pi_j}$; however, it does not satisfy the boundary condition
Eqn.~\ref{eqn:bc} and so it is of limited practical use for estimating gradients.\footnote{Note that since $\bm{\pi}$ is constrained
to lie on the simplex, the relevant velocity fields to consider are defined w.r.t.~an appropriate parameterization like softmax logits
$\bm{\ell}$. It is for these velocity fields that the boundary condition needs to hold and not for $\bm{v}^{\pi_j}$ itself. This is
why Eqn.~\ref{eqn:badvel} can be used for $D=1$, where the boundary condition \emph{does} hold.} 
Intuitively, the problem with Eqn.~\ref{eqn:badvel} is that $\bm{v}^{\pi_j}$ sends mass to infinity.

\section{Adaptive Velocity Fields for the Multivariate Normal Distribution}
\label{sec:avfsupp}

We show that the velocity field
\begin{equation}
\label{eqn:mvnnull}
\bm{\tilde{v}}^{L_{ab}}_{\bm{A}} = \bm{L}\bm{A}^{ab}\bm{L}^{-1} (\bz - \bmu)
\end{equation}
given in the main text is a solution to the corresponding null transport equation.\footnote{This derivation
can also be found in reference \cite{beyond}.}
The transport equation can be written in the form
\begin{equation}
\label{eqn:mvnlogcont}
\frac{\partial}{\partial L_{ab}} \log{q} + \nabla \cdot {\bm{ \tilde{v}}} + {\bm{\tilde{v}}} \cdot \nabla \log q=0
\end{equation}
Transforming to whitened coordinates $\bm{\tilde{z}} = \bm{L}^{-1} (\bm{z}- \bmu)$, the null equation
is given by
\begin{equation}
\nabla \cdot \bm{\tilde{v}} = \bm{\tilde{v}} \cdot \bm{\tilde{z}} 
\end{equation}
We let $\bm{\tilde{v}} = \bm{A}^{ab}\bm{\tilde{z}}$ and compute
\begin{equation}
\nabla_{\bm{\tilde{z}}} \cdot \bm{\tilde{v}} = {\rm Tr} \;\bm{A}^{ab} = 0 = \sum_{ij}  \tilde{z}_i \bm{A}^{ab}_{ij} \tilde{z}_j = \bm{\tilde{v}} \cdot \bm{\tilde{z}} 
\end{equation}
where we have used that $\bm{A}^{ab}$ is antisymmetric. Transforming $\bm{\tilde{v}}$
 back to the given coordinates $\bz$, we
end up with Eqn.~\ref{eqn:mvnnull} (the factor of $\bm{L}$ enters when we transform the vector field).

For the `reference solution' $\bm{v}_0^{L_{ab}}$ we simply use the velocity field furnished by the reparameterization trick, which
is given by
\begin{equation}
(\bm{v}_0^{L_{ab}})_i =  \delta_{ia}  (\bm{L}^{-1} \bm{z})_b 
\end{equation}
Thus the complete specification of $\bm{v}_{\bm{A}}^{L_{ab}}$ is
\begin{equation}
\label{eqn:avfestimator}
(\bm{v}_{\bm{A}}^{L_{ab}})_i =  \delta_{ia}  (\bm{L}^{-1} \bm{z})_b  + (\bm{L}\bm{A}^{ab}\bm{L}^{-1} (\bz - \bmu))_i
\end{equation}
As mentioned in the main text, the computational complexity of using AVF gradients with this class of parameterized velocity fields (including the $\bm{A}$ update equations) is
\begin{equation}
\label{eqn:avfcomp}
\mathcal{O}(D^3 + MD^2)
\end{equation}
This should be compared to the $\mathcal{O}(D^2)$ cost of the reparameterization trick gradient and the $\mathcal{O}(D^3)$
cost of the OMT gradient.
However, the computational complexity in Eqn.~\ref{eqn:avfcomp} is somewhat misleading in that the 
$\mathcal{O}(D^3)$ term arises from matrix
multiplications, which tend to be quite fast. By contrast the OMT gradient estimator involves a singular value decomposition, which
tends to be substantially more expensive than a matrix multiplication on modern hardware. 
In cases where computing the test function involves expensive operations like Cholesky
factorizations, the additional cost reflected in Eqn.~\ref{eqn:avfcomp} is limited (at least for $M \ll D$). For example, as reported in the GP experiment
in Sec.~\ref{sec:co2gp} in the main text where $D=468$, the AVF gradient estimator 
for $M=1$ ($M=5$) requires only $\sim$6\% ($\sim$11\%) more time per iteration.

\subsection{Adaptive Velocity Fields for the Multivariate t-Distribution}

We consider the multivariate t-distribution in $D$ dimensions with probability density
\begin{equation}
\begin{split}
&q_{ \bth}(\bz) = \int_0^\infty q(\tau|\nu) q(\bz | \bm{L}, \tau) d\tau \\
& \propto \frac{1}{|\bm{L}|} \left(1 + \tfrac{1}{\nu}  \bz^{\rm T} \bm{\Sigma}^{-1} \bz\right)^{-\frac{\nu + D}{2}}
\end{split}
\end{equation}
where
\begin{equation}
\nonumber
 q(\tau|\nu) = {\rm Ga}(\tau| \tfrac{\nu}{2},  \tfrac{\nu}{2}) \;\;  \;\; q(\bz | \bm{L}, \tau)=\mathcal{N}(\bz|\bm{0}, \tau^{-\tfrac{1}{2}} \bm{L})
\end{equation}
We want to compute derivatives w.r.t~$L_{ab}$. We compute
\begin{equation}
\begin{split}
&\frac{\partial \log q_{ \bth}(\bz)}{\partial L_{ab}}   = \\
& \frac{\partial}{\partial L_{ab}} \left( -\log |\bm{L}| -\tfrac{\nu + D}{2} \log \left(1 + \tfrac{1}{\nu}  \bz^{\rm T} \bm{\Sigma}^{-1} \bz\right)  \right)  = \\
& - L_{ba}^{-1}  +\tfrac{\nu + D}{\nu} \left(1 + \tfrac{1}{\nu}  \bz^{\rm T} \bm{\Sigma}^{-1} \bz\right)^{-1} (\bm{\Sigma}^{-1} \bz)_a (\bm{L}^{-1} \bz)_b
 \end{split}
\end{equation}
Now suppose  $\bm{v}_{\bm{A}}^{L_{ab}}$ is given as in Eqn.~\ref{eqn:avfestimator}.
Then we have
\begin{equation}
\nabla \cdot \bm{v}_{\bm{A}}^{L_{ab}} = \nabla \cdot \bm{v}_{0}^{L_{ab}} = L_{ba}^{-1}
\end{equation}
and 
\begin{equation}
\nabla q_{ \bth}(\bz) = -\tfrac{\nu + D}{\nu} \left(1 + \tfrac{1}{\nu}  \bz^{\rm T} \bm{\Sigma}^{-1} \bz\right)^{-1}(\bm{\Sigma}^{-1} \bz)
\end{equation}
It is easy to show that 
\begin{equation}
\bm{v}_{\bm{A}}^{L_{ab}} \cdot \nabla q_{ \bth}(\bz) = \bm{v}_{0}^{L_{ab}} \cdot \nabla q_{ \bth}(\bz)
\end{equation}
since the term containing $\bm{A}^{ab}$ vanishes due to the anti-symmetry of $\bm{A}^{ab}$.
Thus one has
\begin{equation}
\bm{v}_{\bm{A}}^{L_{ab}} \cdot \nabla q_{ \bth}=- \tfrac{\nu + D}{\nu} \left(1 + \tfrac{1}{\nu}  \bz^{\rm T} \bm{\Sigma}^{-1} \bz\right)^{-1}(\bm{\Sigma}^{-1} \bz)_a (\bm{L}^{-1} \bz)_b
\end{equation}
Gathering terms we see that $\bm{v}_{\bm{A}}^{L_{ab}}$ satisfies the relevant transport equation, namely
\begin{equation}
\frac{\partial}{\partial L_{ab}} \log{q_\bth} + \nabla \cdot {\bm{v}_{\bm{A}}^{L_{ab}}} + {\bm{v}_{\bm{A}}^{L_{ab}}} \cdot \nabla \log q_\bth=0
\end{equation}
Consequently we have shown that the velocity fields given in Eqn.~\ref{eqn:avfestimator} can be used to a construct adaptive gradient
estimators for the cholesky matrix of the multivariate t-distribution. 

Note that the only property of $q_\bth(\bz)$ that was used in the derivation
was the fact that 
\begin{equation}
\label{eqn:sphsymm}
q_\bth(\bz) \propto \frac{1}{|\bm{L}|} g( \bz^{\rm T} \bm{\Sigma}^{-1} \bz)
\end{equation}
for some scalar density $g(\cdot)$. Thus the velocity fields in Eqn.~\ref{eqn:avfestimator} can in fact be used
to construct adaptive gradient estimators for all distributions of the form given in Eqn.~\ref{eqn:sphsymm}, i.e.~for all
elliptical distributions.

\section{Mixture distributions}
\label{sec:mixsupp}

\begin{table*}[h]
\begin{center}
    \begin{tabu}{|c|c|c|c|}    \hline
     Distribution Name \cellcolor[gray]{0.90} & \cellcolor[gray]{0.90} Component Distributions & \cellcolor[gray]{0.90} Velocity Field
     & \cellcolor[gray]{0.90} Computational Complexity \\  \tabucline[1pt]{-}
    DiagNormalsSharedCovariance & $\mathcal{N}(\bz | \bmu_j, \bm{\sigma})$ & 
                               Eqn.~\ref{eqn:arbmumixsoln} & $\mathcal{O}(K^2 D)$ \\ \hline
    ZeroMeanGSM  & $\mathcal{N}(\bz | \bm{0}, \lambda_j \bm{\sigma})$  & Eqn.~\ref{eqn:gsmsoln2} & $\mathcal{O}(K D) $ \\ \hline
    GSM  & $\mathcal{N}(\bz | \bmu_j, \lambda_j \bm{\sigma})$  & Eqn.~\ref{eqn:gengsm}& $\mathcal{O}(K^2 D)$  \\ \hline
    DiagNormals  & $\mathcal{N}(\bz | \bmu_j, \bm{\sigma}_j)$  & Eqn.~\ref{eqn:arbmixsoln2} & $\mathcal{O}(K D)$ \\ \hline
    \end{tabu}
\end{center}  
\caption{Four families of mixture distributions for which we can compute pathwise derivatives. The names are those used in Fig.~1b in the main text.}
 \label{tab:mix}
\end{table*}

In Table \ref{tab:mix} we summarize the four families of mixture distributions for which we have found closed form solutions to the 
transport equation. The first one was presented in the main text; here we also present the solutions for the three other families of mixture distributions.

\subsection{Pairwise Mass Transport}

We begin with the transport equation for $\pi_j$, which reads
\begin{equation}
q_{ \bth_j} + \nabla_\bz \cdot \left( q_\bth \bm{v}^{\pi_j} \right) = 0
\end{equation}
Introducing softmax logits $\ell_j$ given by
\begin{equation}
\pi_j = \frac{e^{\ell_j}}{\sum_k e^{\ell_k}}
\end{equation}
and using the fact that
\begin{equation}
\frac{\partial \pi_k}{\partial \ell_j} = \pi_j (\delta_{kj} - \pi_k)
\end{equation}
we observe that the velocity field for $\ell_j$ satisfies the following transport equation
\begin{equation}
\label{eqn:logittransport}
\pi_j\left(q_{\bth_j} - \sum_{k} \pi_k q_{\bth_k}\right) + \nabla_\bz \cdot \left( q_\bth \bm{v}^{\ell_j} \right) = 0
\end{equation}
We substitute Eqn.~\ref{eqn:vlogits} for $\bm{v}^{\ell_j}$ and compute the divergence term, which yields
\begin{equation}
\nonumber
\begin{split}
&\nabla_\bz \cdot \left( q_\bth \bm{v}^{\ell_j} \right) =
\nabla_\bz \cdot \left( q_\bth \pi_j \sum_{k\ne j} \pi_k \tilde{\bm{v}}^{jk} \right)  = \\
& - \pi_j \sum_{k\ne j} \pi_k  \left( q_{\bth_j} - q_{\bth_k} \right) = 
- \pi_j \left(q_{\bth_j} - \sum_{k} \pi_k q_{\bth_k} \right) 
\end{split}
\end{equation}
Thus $\bm{v}^{\ell_j}$ satisfies the relevant transport equation Eqn.~\ref{eqn:logittransport}.

\subsection{Zero Mean Discrete Gaussian Scale Mixture}
\label{sec:scalemix}

Here each component distribution is specified by $q_{\bth_j}(\bz) = \mathcal{N}(\bz | \bm{0}, \lambda_j \bm{\sigma})$,
where each $\lambda_j$ is a positive scalar.
Defining $\tilde{\bz} = \bz \odot \bm{\sigma}^{-1}$ and making use of radial coordinates with $r = || \tilde{\bz} ||$
we find that a solution of the form in Eqn.~\ref{eqn:vlogits} reduces to
\begin{equation}
\label{eqn:gsmsoln2}
\bm{v}^{\ell_j} = \pi_j {\rm diag}(\bm{\sigma})  \left( \tilde{\bm{v}}^{j} -  \sum_{k} \pi_k \tilde{\bm{v}}^{k} \right)
\end{equation}
where
\begin{equation}
\label{eqn:gsmsoln}
\begin{split}
 &\tilde{\bm{v}}^{j} = \frac{\tilde{\Phi}(\frac{r}{ \lambda_j})} {q_\bth \lambda_j^{D-1} \prod_{i=1}^D \sigma_i  } \hat{\bm{r}} \qquad {\rm and} \\
 &\tilde{\Phi}(z) = \frac{z^{1-D}}{(2\pi)^{D/2}} \int_{z}^{\infty}  \tilde{z}^{D-1} e^{-\tilde{z}^2/2}d\tilde{z}
 \end{split}
\end{equation}
The `radial CDF' $\tilde{\Phi}$ in Eqn.~\ref{eqn:gsmsoln} can be computed analytically.
 In even dimensions we find\footnote{The notation $k!!$ refers to the double
factorial of $k$, which occurs in this context through the identity $(2n-1)!! = 2^n \Gamma(n+\tfrac{1}{2}) / \sqrt{\pi}$.}
\begin{equation}
\tilde{\Phi}(z) = \frac{e^{-z^2/2}}{(2\pi)^{D/2}} \sum_{k=0}^{\tfrac{D}{2}-1} \frac{(D-2)!!}{(2k)!!} z^{2k + 1 - D}
\end{equation}
and in odd dimensions we find
\begin{equation}
\begin{split}
&\tilde{\Phi}(z) = \frac{e^{-z^2/2}}{(2\pi)^{D/2}}   \sum_{k=1}^{\tfrac{D-1}{2}} \frac{(D-2)!!}{(2k -1)!!} z^{2k-D}  + \\
&(D-2)!!\sqrt{\tfrac{\pi}{2}}\frac{1\!-\! {\rm erf}(\tfrac{z}{\sqrt{2}})}{(2\pi)^{D/2}z^{D-1}}
\end{split}
\end{equation}
where ${\rm erf}(\cdot)$ is the error function. Note that in contrast to all the other solutions in Table \ref{tab:mix},
 Eqn.~\ref{eqn:gsmsoln2} for $D$ even does not involve any error functions. 

We show explicitly that Eqn.~\ref{eqn:gsmsoln2} is a solution of the corresponding transport equation. The derivations
for the other families of mixture distributions are similar. It is enough to show that
\begin{equation}
\sum_i \frac{\partial}{\partial z_i} \left( q_{\bth} \tilde{\bm{v}}^{j}_i  \right) =
 \sum_i \frac{\partial}{\partial z_i} \left(\frac{\tilde{\Phi}(\frac{r}{ \lambda_j})} { \lambda_j^{D-1} \prod_{i=1}^D \sigma_i  } \hat{\bm{r}}_i\right)= 
- q_{\bth_j}(\bz)
\end{equation}
Using the identities
\begin{equation}
\frac{\partial r}{\partial z_i} = \frac{z_i}{r \sigma_i^2} \qquad \qquad
\frac{\partial}{\partial z_i} = \frac{\partial r}{\partial z_i} \frac{\partial }{\partial r} \qquad\qquad
 \hat{\bm{r}}_i = \frac{z_i}{r}
\end{equation}
which follow from the definition $r^2 = \sum_i  \frac{z_i^2}{ \sigma_i^2}$, we have
\begin{equation}
\begin{split}
&\sum_i \frac{\partial}{\partial z_i} \left(\tilde{\Phi}(\tfrac{r}{ \lambda_j})  \hat{\bm{r}}_i\right) =
\frac{1}{\lambda_j}\tilde{\Phi}^\prime(\tfrac{r}{ \lambda_j}) \sum_i   \frac{z_i^2}{ \sigma_i^2 r^2} = \\
&\frac{1}{\lambda_j}\tilde{\Phi}^\prime(\tfrac{r}{ \lambda_j}) \frac{r^2}{r^2} = 
\frac{1}{\lambda_j}\tilde{\Phi}^\prime(\tfrac{r}{ \lambda_j}) 
\end{split}
\end{equation}
By construction we have that
\begin{equation}
{\Phi}^\prime(\tfrac{r}{ \lambda_j}) = - \frac{1}{(2\pi)^{D/2}}  e^{-\tfrac{1}{2}r^2/\lambda_j^2} = 
-\left( \lambda_j^{D} \prod_{i=1}^D \sigma_i\right) q_{\bth_j}(\bz)
\end{equation}
Comparing terms, we see that $\tilde{\bm{v}}^{j}$ is indeed a solution of the transport equation for $\pi_j$ as desired.

\subsection{Mixture of Diagonal Normal Distributions}
\label{sec:scalemixii}

Here each component distribution is given by
\begin{equation}
q_{\bth_j}(\bz) = \mathcal{N}(\bz | \bmu_j, \bm{\sigma}_j) \qquad {\rm for} \;\; j=1,2,...,K
\end{equation}
For $i=1,...,D$ define 
\begin{equation}
\tilde{z}_{ji} = \frac{z_i - \mu_{i}}{\sigma_{ji}} \qquad
\bar{z}_{ji} = \frac{z_i - \mu_{i}}{\sigma_i^0} \qquad
r_{ji}^2 = \sum_{k < i} \tilde{z}_{jk}^2 + \sum_{k > i} \bar{z}_{jk}^2
\end{equation}
where $\bm{\sigma}^0$ is an arbitrary reference scale.
We find that if we define\footnote{Here we take the empty products $\prod_{k < 1}$ and $\prod_{k > D}$ to be equal to unity.}
\begin{equation}
\label{eqn:arbmixsoln}
 \breve{\bm{v}}^{j} = \sum_i \frac{\left(\Phi(\tilde{z}_{ji} ) - \Phi(\bar{z}_{ji}) \right)
 \phi(r_{ji}^2)}{q_\bth \prod_{k < i} \sigma_{jk} \prod_{k > i} \sigma_{k}^0}  \hat{\bz}_i
 \end{equation}
then we can construct a solution of the form specified in Eqn.~\ref{eqn:vlogits} that is given by
\begin{equation}
\label{eqn:arbmixsoln2}
\bm{v}^{\ell_j} = \pi_j\left( \breve{\bm{v}}^{j} -  \sum_{k} \pi_k \breve{\bm{v}}^{k} \right)
\end{equation}
Since the reference scale $\bm{\sigma}^0$  is {\it arbitrary}, this is actually a parameterized family of solutions. Thus
this solution is in principle amenable to the Adaptive Velocity Field approach described in Sec.~\ref{sec:avf} in the main text. In
addition, the ordering of the dimensions $i=1, ..., D$ that occurs implicitly in the telescopic structure of Eqn.~\ref{eqn:arbmixsoln}
is also arbitrary. Thus Eqn.~\ref{eqn:arbmixsoln} corresponds to a very large family of solutions. In practice we use
the fixed ordering given in Eqn.~\ref{eqn:arbmixsoln} and choose 
\begin{equation}
{\sigma}^0_i \equiv \min\limits_{j \in [1, K]} \sigma_{ji}
\end{equation}
We find this works pretty well empirically.

\subsection{Discrete GSM}
\label{sec:partiallysharedcov}

Here each component distribution is given by
\begin{equation}
q_{\bth_j}(\bz) = \mathcal{N}(\bz | \bmu_j, \lambda_j \bm{\sigma}) \qquad {\rm for} \;\; j=1,2,...,K
\end{equation}
where each $\lambda_j$ is a positive scalar. We can solve the corresponding 
transport equation for the mixture weights by superimposing the solutions in Eqn.~\ref{eqn:arbmumixsoln}
and Eqn.~\ref{eqn:gsmsoln2}. In more detail, in this case the solution to Eqn.~\ref{eqn:pairwisetransport}
is given by 
\begin{equation}
\label{eqn:gengsm}
\bar{\bm{v}}^{jk}  = \underbrace{\tilde{\bm{v}}^{jk;\lambda=\lambda_0}}_{{\textrm{soln.~from Eqn.~}\ref{eqn:arbmumixsoln}} }
 \! \! \! \! \! +\tilde{\bm{w}}^{j} - \tilde{\bm{w}}^{k} 
\end{equation}
where 
\begin{equation}
\tilde{\bm{w}}^{j} = \underbrace{ \tilde{\bm{v}}^{j;\lambda_j} -  \tilde{\bm{v}}^{j;\lambda=\lambda_0}}_{{\textrm{solutions from Eqn.~}\ref{eqn:gsmsoln2}}}
\end{equation}
Analogously to the reference scale $\bm{\sigma}^0$ in Sec.~\ref{sec:scalemixii}, $\lambda_0$ is arbitrary. As
such Eqn.~\ref{eqn:gengsm} is a parametric family of solutions that is amenable to the Adaptive Velocity Field approach.
Intuitively, we use the solutions from Eqn.~\ref{eqn:arbmumixsoln} to effect mass transport between component means and solutions
from Eqn.~\ref{eqn:gsmsoln2} to shrink/dilate covariances.

\subsection{Mixture of Multivariate Normals with Shared Diagonal Covariance}

To finish specifying the solution Eqn.~\ref{eqn:arbmumixsoln} given in the main text, 
we define the following coordinates:
\begin{equation}
\begin{split}
\nonumber
\tilde{\bz} &= \bz \odot \bm{\sigma}^{-1} \qquad \tilde{\bmu}^j = \bmu^j \odot \bm{\sigma}^{-1}  \qquad
\hat{\bmu}^{jk} = \frac{\tilde{\bmu}^j-\tilde{\bmu}^k}{||\tilde{\bmu}^j-\tilde{\bmu}^k||} \\ 
\tilde{\mu}^{jk}_\parallel &= \tilde{\bmu}^{j} \cdot \hat{\bmu}^{jk}  \qquad
\tilde{z}_\parallel^{jk} = \tilde{\bz} \cdot \hat{\bmu}^{jk} \qquad 
\tilde{\bz}_\bot^{jk} = \tilde{\bz} - \tilde{z}_\parallel^{jk} \hat{\bmu}^{jk}
\end{split}
\end{equation}

\subsection{Velocity Fields for the Component Parameters of Multivariate Mixtures}

Suppose $\bm{v}^{\bth_j}_{\rm{single}}$ is a solution of the single-component transport equation for
the parameter $\bth_j$, i.e.~
\begin{equation}
\frac{\partial q_{\bth_j}}{\partial \bth_j} + \nabla \cdot(q_{\bth_j} \bm{v}^{\bth_j}_{\rm{single}}) = 0
\end{equation}
Then 
\begin{equation}
\bm{v}^{\bth_j} = \frac{ \pi_j q_{\bth_j}} {q_\bth} \bm{v}^{\bth_j}_{\rm{single}}
\end{equation}
is a solution of the multi-component transport equation, since
\begin{equation}
\begin{split}
& \frac{\partial q_\bth(\bm{z})}{\partial \bth_j} + \nabla \cdot(q_\bth(\bm{z}) \bm{v}^{\bth_j} )  = 
  \pi_j \frac{\partial q_{\bth_j}(\bm{z})}{\partial \bth_j} + \nabla \cdot(q_\bth(\bm{z}) \bm{v}^{\bth_j} ) \\ &=
    \pi_j \left( \frac{\partial q_{\bth_j}(\bm{z})}{\partial \bth_j} + \nabla \cdot(q_{\bth_j}(\bm{z}) \bm{v}^{\bth_j}_{\rm{single}} ) \right) = 0
 \end{split}
\end{equation}
This completes the derivation for the claim about $\bm{v}^{\bth_j}$ made at the beginning of Sec.~\ref{sec:mix} in the main text.

\subsection{Pairwise Mass Transport and Control Variates}
For $j=1,...,K$ define $K \times K$ square matrices $A_{ik}^j$ such that all the rows and columns of each $A_{ik}^j$ sum to zero.
Then
\begin{equation}
\bm{w}^{\ell_j} = \sum_{i, k} A_{ik}^j \tilde{\bm{v}}^{ik} 
\end{equation}
is a null solution to the transport equation for $\bm{v}^{\ell_j}$, Eqn.~\ref{eqn:logittransport}. While we have not done so ourselves, these null solutions could be used to adaptively move mass among the $K$ component distributions of a mixture instead of using the recipe in Eqn.~\ref{eqn:vlogits}, which takes mass from each component
distribution $j$ in proportion to its mass $\pi_j$ (which is in general suboptimal).

\section{Experimental Details}
\label{sec:expsupp}

\subsection{Synthetic Test Function Experiments}

We describe the setup for the experiments in Sec.~\ref{sec:synthavf} and Sec.~\ref{sec:synthmix} in the main text.

For the experiment in Sec.~\ref{sec:synthavf} the dimension is fixed to $D=50$ and the mean of $q_\bth$ 
is fixed to the zero vector. The Cholesky factor $\bm{L}$ that
enters into $q_\bth$ is constructed as follows. The diagonal of $\bm{L}$ consists of all ones. To construct the off-diagonal
terms we proceed as follows. We populate the entries below the diagonal of a matrix $\Delta \bm{L}$ by drawing each
entry from the uniform distribution on the unit interval. Then we define $\bm{L} = \mathbb{1}_D + r \Delta \bm{L}$. Here $r$
controls the magnitude of off-diagonal terms of $\bm{L}$ and appears on the horizontal axis of Fig.~\ref{fig:mvnreach} in the main text.
The three test functions are constructed as follows. First we construct a strictly lower diagonal matrix $ \bm{Q}^\prime$ by
drawing each entry from a bernoulli distribution with probability 0.5. We then define $ \bm{Q}= \bm{Q}^\prime + \bm{Q}^{\prime T}$.
The cosine test function is then given by
\begin{equation}
f(\bz) = \cos \left(\sum_{i, j} Q_{ij} z_i / D \right)
\end{equation}
The quadratic test function is given by
\begin{equation}
f(\bz) = \bz^{T} \bm{Q} \bz
\end{equation}
The quartic test function is given by
\begin{equation}
f(\bz) = \left(\bz^{T} \bm{Q}\bz\right)^2 
\end{equation}
The AVF gradient uses $M=1$ and we train $\bm{A}^{ab}$ to (approximate) convergence before estimating the gradient variance.

For the experiment in Sec.~\ref{sec:synthmix} that is depicted in Fig.~\ref{fig:mixreach} the test function is fixed to $f(\bz) = ||\bz||^2$. The distributions $q_\bth$ are constructed as follows. For the distributions that 
admit a parameter $\bmu_j$, each $\bmu_j$ is sampled from the sphere centered at $\bz=\bm{0}$ with
radius 2. For the distribution whose velocity field is given in Eqn.~\ref{eqn:gsmsoln2}, the mean is fixed to $\bm{0}$. The covariance 
matrices are sampled from a narrow distribution centered at the identity matrix. Consequently the different mixture components have little overlap.

For the experiment in Sec.~\ref{sec:synthmix} that is depicted in Fig.~\ref{fig:mixreach2} the test function is also fixed to $f(\bz) = ||\bz||^2$. The distributions $q_\bth$ are constructed as follows. The $K$ means are placed uniformly around the unit circle. The covariance of each component distribution is given by $\sigma^2 \mathbb{1}$, where $\sigma$ is the parameter that is varied along the horizontal axis of the figure. For the gradient estimator derived from the transport equation, we use the estimator described in Sec.~\ref{sec:scalemixii}, although in this particular case the estimator given by Eqn.~\ref{eqn:arbmumixsoln} yields identical results.

In all cases the gradients can be computed analytically, which makes it easier to reliably estimate the variance of the gradient
estimators. 

\subsection{Gaussian Process Regression}

We use the Adam optimizer \cite{kingma2014adam} to optimize the ELBO with single-sample gradient estimates. We chose
the Adam hyperparameters by doing a grid search over the learning rate and $\beta_1$. For the AVF gradient estimator 
the learning rate and $\beta_1$ are allowed to differ between $\bth$ and $\bm{\lambda}$ gradient steps (the latter needs a much larger learning rate for good results). 
For each combination of hyperparameters we
did 500 training iterations for five trials with different random seeds and then chose the combination that yielded the highest mean ELBO after 500 iterations. 
We then trained the model for 500 iterations, initializing with another random number seed. The figure in the main text shows the training curves for that single run.
We confirmed that other random number seeds give similar results.

\subsection{Baseball Experiment}
There are 18 baseball players and the data consists of 45 hits/misses for each player. The model has two global 
latent random variables, $\phi$ and $\kappa$, with priors
$\rm{Uniform}(0,1)$ and $\rm{Pareto}(1, 1.5) \propto \kappa^{-5/2}$, respectively. There are 18 local latent random variables, $\theta_i$ for $i=0,...,17$, with
$p(\theta_i) = \rm{Beta}(\theta_i | \alpha= \phi  \kappa, \beta=(1 - \phi)  \kappa)$. The data likelihood factorizes into 45 Bernoulli observations with mean chance of success $\theta_i$ for each player $i$. All variational approximations are formed in the unconstrained space
 $\{\rm{logit}(\phi), \log(\kappa - 1), \rm{logit}(\theta_i)\}$. The mean field variational approximation consists of a diagonal Normal distribution in the unconstrained space, while the mixture variational approximation consists of $K$ diagonal Normal distributions in the same space. We use the Adam optimizer for training with a learning rate of $5 \times 10^{-3}$ \cite{kingma2014adam}. 

\subsection{Continuous State Space Model}

We consider the following simple model with two dimensional observations $\bx_t$ and two dimensional
latent random variables $\bz_t$: 
\begin{equation}
p(\bx_{1:T}, \bz_{1:T}) = p(\bz_1) p(\bx_1 | \bz_1) \prod_{t=2}^T p(\bz_t | \bz_{t-1} ) p(\bx_t | \bz_t)
\end{equation}
where 
\begin{equation}
\begin{split}
&p(\bz_1) = \mathcal{N}(\bz_1 | \bm{0}, \sigma_z \mathbb{1}_2)  \\
&p(\bz_t | \bz_{t-1} ) = \mathcal{N}(\bz_t  | \bm{T}  \bz_{t-1},  \sigma_z \mathbb{1}_2)  \\
&p(\bx_t | \bz_t) = \mathcal{N}(\bx_t | \bm{\mu}(\bz_t), \sigma_x \mathbb{1}_2)
\end{split}
\end{equation}
and 
\begin{equation}
\begin{split}
&\bm{\mu}(\bz_t) = (z_{1t}^2, 2z_{2t}) \qquad
\sigma_z=\frac{1}{2} \qquad
\sigma_x=\frac{1}{4} \\
&\bm{T} =  \frac{1}{2}\begin{pmatrix}
    -\sin(\theta) &  \cos(\theta)  \\
    \cos(\theta) & \sin(\theta) 
  \end{pmatrix} \qquad
  \theta = \frac{\pi}{4} 
  \end{split}
\end{equation}
The quadratic term $z_{1t}^2$ in $\bm{\mu}(\bz_t)$ results in a highly multi-modal posterior. We generate 1000 sequences with $T=10$ and 
use 800 for training and 200 for testing.  The model dynamics are assumed to be known, and
 the variational family is constructed along the lines of the DKS inference network
in \cite{krishnan2017structured}.
We use the pathwise gradient estimators introduced in Sec.~\ref{sec:mix} in the main text.

\subsection{Deep Markov Model}

The training data consist of 229 sequences with a mean length of 60 time steps from the JSB Chorales polyphonic music dataset considered
in \cite{boulanger2012modeling}. Each time slice in a sequence spans a quarter note and is represented by an 88-dimensional binary vector.
We use a Bernoulli likelihood. The dimension of the latent $\bz_t$ at each time step is 32.
The inference network a.k.a.~variational family follows the DKS variant described
in \cite{krishnan2017structured}.
Similarly, the architecture of the various neural network components follows the architectures used in \cite{krishnan2017structured}.
In particular the RNN dimension is fixed to be 600 and the dimension of the hidden layer in the neural network that
parameterizes $p(\bz_t | \bz_{t-1})$ is 200; all other neural network hidden layers are 100-dimensional. 
We use a mini-batch size of 20.  Following \cite{krishnan2017structured} we anneal the contribution of KL divergence-like terms over the course of optimization (we use a linear schedule). 
We use the Adam optimizer \cite{kingma2014adam} with gradient clipping and an exponentially decaying learning rate and do up to 7000 epochs of learning. We do a grid search over optimization hyperparameters, which include the learning rate, $\beta_1$, the KL annealing schedule, and temperature (the latter only in the case of the Gumbel Softmax estimators). We use the validation set to fix the hyperparameters and then report results on the test set. The reported test
ELBOs use a 200-sample estimator and are normalized per timestep.

\subsubsection{\bf{Gradient Estimators}}
\label{sec:gumb}

For $K=2$ the mixture distributions have arbitrary diagonal covariance matrices; consequently the pathwise gradient estimator is of the form described in Sec.~\ref{sec:scalemixii}. For $K=3$ the mixture distributions share diagonal covariance matrices at each time step (we make this choice to limit the total number of parameters); 
consequently the pathwise gradient estimator is of the form described in Sec.~\ref{sec:sharedcov} in the main text. 

The two variants of the Gumbel Softmax estimator we use are more similar to the approach adopted in \cite{jang2016categorical} than to
the approach adopted in \cite{maddison2016concrete}. In particular we do {\it not} relax the objective function in the manner of \cite{maddison2016concrete} so that the resulting gradient estimators are biased. In GS-Soft we draw a $K$ dimensional
vector $\bm{y}$ from the Gumbel Softmax distribution so that $\bm{y}$ lies in the interior of the $K-1$ dimensional simplex. To
generate a sample from the mixture $q(\bz_t | \cdot)$ we draw a $D$-dimensional sample $\bm{\epsilon} \sim \mathcal{N}(\bm{0},\bm{1})$ and form 
the sample $\bz_t$ via
\begin{equation}
\label{eqn:onefellswoop}
z_{t,i} = \sum_{k=1}^K y_k \left( \mu_{ki} + \epsilon_i \sigma_{ki} \right) \qquad {\rm for} \qquad i=1,...,D
\end{equation}
In GS-Hard we adopt the same approach as in GS-Soft, except $\bm{y}$ is discretized via arg max, c.f. the straight-through
estimator in \cite{jang2016categorical}.
We adopt the approach in Eqn.~\ref{eqn:onefellswoop} so that we do not need to introduce variational distributions of the 
form $q(\bm{\pi}_t | \cdot)$, as we expect that this additional variational relaxation would lead to looser ELBO bounds (and would
make a direct comparison to variational setups without ELBO terms of the form $\log q(\bm{\pi}_t | \cdot)$ more difficult). 

We do not report numbers for the score function estimator, since the extremely
 high variance---$\mathcal{O}(10^5)$ times higher than for the
pathwise gradient estimator---prevented us from obtaining competitive results. In particular we were unable to obtain test ELBOs
above -9.0 nats; by contrast a mean field variational family with diagonal Normal distributions of the
form $q(\bz_t | \bx_t)$ at each time step can achieve $\sim-8.0$ nats.

\subsection{VAE}

We train using MNIST 50k and fix the latent dimensionality to 50. 
The prior is Normal and the likelihood is Bernoulli. 
We fix the number of hidden units in the inference network to 400 and the number of hidden units in the decoder network to 200.
We use the Adam optimizer and do a grid search over the following optimization
hyperparameters: learning rate, $\beta_1$, and the temperature (for the Gumbel Softmax estimators). 
We train all models for 3000 epochs with a batch size of 256. Test ELBOs are computed with a 50-sample estimator. For details
on the Gumbel Softmax estimators, see Sec.~\ref{sec:gumb}.


\end{document}